\documentclass[lettersize,journal]{IEEEtran}
\usepackage{amsmath,amsfonts,amssymb}
\usepackage{array}
\usepackage[caption=false,font=normalsize,labelfont=sf,textfont=sf]{subfig}
\usepackage[font=footnotesize,labelfont=bf]{caption}
\usepackage{xcolor}

\usepackage{colortbl}
\usepackage{arydshln}
\usepackage{booktabs}
\usepackage{footnote}
\usepackage[flushleft]{threeparttable}

\usepackage{textcomp}
\usepackage{url}
\usepackage{verbatim}
\usepackage{graphicx}
\usepackage{cite}
\usepackage{hyperref}
\usepackage{tabularx}
\usepackage{color}
\usepackage{graphicx}
\usepackage{textcomp}
\usepackage{wrapfig}
\usepackage[tableposition = top]{caption}
\usepackage{booktabs}%提供命令\toprule、\midrule、\bottomrule
\usepackage{siunitx}

\usepackage{multirow}%提供跨列命令\multicolumn{}{}{}
\usepackage{algpseudocode}
\usepackage[ruled,vlined]{algorithm2e}

\usepackage{floatrow}
\newfloatcommand{capbtabbox}{table}[][\FBwidth]
\usepackage{blindtext}

\floatstyle{plaintop}
\restylefloat{table}

\usepackage[T1]{fontenc}
\usepackage[utf8]{inputenc}
\usepackage{babel}

\usepackage{subfig}% http://ctan.org/pkg/subfig
\usepackage{booktabs}% http://ctan.org/pkg/booktabs

\usepackage{graphicx}
\usepackage{caption}
\usepackage{lipsum}

\usepackage{soul}

\begin{document}

% \title{A Long Horizon Planning Framework for Rearranging Deformable Linear Objects under Reachability and Contact Constraints}

% \title{A Self-supervised Bimanual Manipulation Framework for Rearranging Deformable Linear Objects under Reachability and Contact Constraints}

% \title{Rearranging Deformable Linear Objects with Bimanual Manipulation under Constraints of Reachability and Obstacles}

% \title{Interleaving Planning and Control for Constrained Bimanual Manipulation of Deformable Linear Objects without Goal Specification
% }

% \title{
% Towards Flexible Constrained Bimanual Manipulation of Deformable Linear Objects Interleaving Planning and Control
% }

% \title{
% A Planning and Control Framework for Flexible Constrained Bimanual Manipulahttps://www.overleaf.com/projecttion without Strong Assumptions
% }

% \title{
% A Long-Horizon Framework for Dexterous Manipulation under Environmental and Reachability Constraints using Contrastive Planning
% }

\title{
A Dual-Arm Collaborative Framework for Dexterous Manipulation in Unstructured Environments with Contrastive Planning
}
% {
% Rearranging Deformable Linear Objects with Bimanual Manipulation under Constraints of Reachability and Obstacles}

\author{Shengzeng Huo, Fangyuan Wang, Luyin Hu, Peng Zhou, Jihong Zhu, Hesheng Wang and David Navarro-Alarcon
        % <-this % stops a space
\thanks{This work is supported by the Research Grants Council (RGC) of Hong Kong under grants 14203917 and 15212721.}% <-this % stops a space
\thanks{S. Huo, F. Wang, L. Hu, P. Zhou and D. Navarro-Alarcon are with The Hong Kong Polytechnic University, Department of Mechanical Engineering, KLN, Hong Kong.}
\thanks{J. Zhu is with the University of York, School of Physics, Engineering and Technology, UK.}
\thanks{H. Wang is with the Shanghai Jiatong University, Department of Automation, Shanghai, China.}

}

\markboth{IEEE/ASME Transactions on Mechatronics}%
{Huo \MakeLowercase{\textit{et al.}}:  }


\maketitle

\begin{abstract}
Most object manipulation strategies for robots are based on the assumption that the object is rigid (i.e., with fixed geometry) and the goal's details have been fully specified (e.g., the exact target pose).
However, there are many 
% symbolic 
tasks that involve spatial relations in human environments where these conditions may be hard to satisfy, e.g., bending and placing a cable inside an unknown container.
To develop advanced robotic manipulation capabilities in unstructured environments that avoid these assumptions, we propose a novel long-horizon framework that exploits contrastive planning in finding promising collaborative actions.
% To develop advanced manipulation capabilities, we present a novel framework for solving long-horizon rearranging task that avoid these assumptions.
Using simulation data collected by random actions, we learn an embedding model in a contrastive manner that encodes the spatio-temporal information from successful experiences, which facilitates the subgoal planning through clustering in the latent space.
Based on the keypoint correspondence-based action parameterization,
% Formulating the manipulation as keypoint correspondence-based alignment between two states, 
we design a leader-follower control scheme for the collaboration between dual arms.
All models of our policy are automatically trained in simulation and can be directly transferred to real-world environments.
% To validate the proposed framework, we conduct a detailed experimental study with multiple bimanual manipulation tasks in complex configurations and subject to environmental and reachability constraints.
To validate the proposed framework, we conduct a detailed experimental study on a complex scenario subject to environmental and reachability constraints in both simulation and real environments.
\end{abstract}


\section{Introduction}
% Bimanual manipulation provides more inputs.
\IEEEPARstart{B}{imanual} manipulation allows to perform more dexterous behaviors than what single-arm systems can do \cite{9788476}; The availability of an additional arm enables robots to perform various complex long-horizon tasks, i.e., those that require to perform several multi-step subtasks over a long time sequence; Examples of these challenging tasks include assembling furniture \cite{FanXie2020DeepIL}, 
spreading a tablecloth \cite{8957044}, grasping and opening a bottle \cite{Chitnis2020EfficientBM}, etc.
% Although great success has been achieved in robotic manipulation in recent years, most existing bimanual strategies are only designed for single action tasks. 
Due to the drastic increase in planning complexity of long-horizon manipulation, the majority of methods further assume rigidity of the manipulated objects and a fully specified goal (e.g. the exact target pose) \cite{Simeonov2020ALH}. 
However, these assumptions are hard to satisfy in some real-world scenarios. 
For example, the case where a robot is commanded to pick a deformable cable from a cluttered environment and arrange it inside a box; The relative spatial relationship ``inside the box'' represents the desired goal rather than the specific target pose or shape of the cable \cite{8106734}.  
Our aim in this paper is precisely to develop methods that break these two assumptions and thus make robot manipulation more applicable for real-world dynamic scenarios, where environmental and reachability constraints are ubiquitous.

Compared with rigid objects, manipulating deformable objects is much more challenging due to their complex mechanical structure (i.e., variable morphology and the high number of degrees of freedom).
Although great success has been achieved in managing the high-dimensional configuration of these types of objects with dual arms (e.g. \cite{navarro2016automatic, yu2022shape, zhu2021vision}), shaping deformable objects without a goal specification remains an open research problem.
In this paper, we provide a solution to this problem in the context of automatically rearranging a deformable linear object (DLO) in a planar setting while simultaneously satisfying geometric constraints.
We choose this case of study since it can be considered a prerequisite for deformable object manipulation (DOM) tasks with fixed contacts \cite{Mitrano2021LearningWT,Wang2022OfflineOnlineLO,zhu2018dual}.
There are several challenges in this setting: (1) Lack of a goal specification; (2) Nonlinear dynamics of the system in unstructured environments; (3) Long-horizon planning complexity; (4) High-dimensional continuous state-action spaces.% of the DLOs and dual arms.

Many researchers have previously addressed the robotic manipulation of soft materials, see \cite{9721534} for a recent review. 
However, most existing methods (either model-based \cite{Zhang2022LearningGM} or model-free \cite{8106734,9634846}) only consider simple tasks that require few steps. 
To deal with the complexity of planning long-horizon tasks with DOM, some works have formulated it as a multistep decision-making problem \cite{yin2021modeling}. 
Point-pair correspondences are utilized in \cite{sundaresan2020learning} for goal-conditioned control, which requires several intermediate subgoals provided by human demonstration. 
The planner in \cite{Lippi2020LatentSR} exploits an encoder-decoder structure to deal with the high dimensionality of the captured visual observations.
However, its generalizability relies on the richness of the collected data.
The method in \cite{WilsonYan2020LearningPR} uses the learned object dynamics to implement a simplified version of a Model Predictive Control (MPC) for DOM. However, these types of methods are not able to handle tasks without a complete goal specification.

% Based on Visual Foresight framework \cite{Finn2017DeepVF}, \cite{Hoque2022VisuoSpatialFF} utilizes cross entropy method to plan action sequences to minimize the performance cost. However, it requires high similarity simulator with the real manipulation environment and huge data collection with random action.  
 
% \textbf{Manipulation in Unstructured Environment} 
DOM in unstructured environments is difficult since the actions are constrained and the physical dynamics are complex \cite{Mitrano2021LearningWT}. There are some works that attempt to solve this challenge without learning the dynamics. However, the majority of them adopt ad-hoc solutions, such as a customized gripper with fixed contacts in \cite{zhu2019robotic}, task-related action primitives in \cite{Huo2022KeypointBasedPB} and simplified state representation with markers in \cite{mcconachie2020manipulating}.
% exploits environmental contacts to manipulate a DLO with assumptions of fixed contact between the end-effector and the object. 
% Although DOM with non-fixed contact is solved in \cite{Huo2022KeypointBasedPB}, it rely on task-related knowledge to design action primitives.
% By interleaving global planning and local control, the efficiency of \cite{mcconachie2020manipulating} is low owing to their sampling-based methods.
Other researchers leverage on data-driven methods \cite{zhou2021lasesom} to avoid modeling the dynamics explicitly. 
% Since collecting real-world data is time-consuming, most of them learn the policy in simulation and then transfer it to experimental platforms.
\cite{Mcconachie2018EstimatingMU} estimates the utility of multiple alternative dynamic models for model-based control.
% By formulating DOM as a multiarmed bandit problem, the policy in \cite{Mcconachie2018EstimatingMU} explores with high-quality models through estimating their utility.  
The method in \cite{Mitrano2021LearningWT} tries to recover from unreliable situations that the unconstrained dynamic model fails.
\cite{matas2018sim} adopts domain randomization techniques to adapt the policy learned in simulation to real situations.
However, most of these methods require a great number of resources to learn the complex dynamics in simulation and their performance are affected when they are applied to reality.
% Without being highly dependent on the dynamics of the object, our proposed planning and control framework enables the trained policy to be transferred from simulation to real-world without fine-tuning.

To provide a feasible solution to these problems, in this paper we present a novel algorithm for long-horizon bimanual manipulation. 
In contrast with most algorithms in the literature, our approach does not rely on object rigidity and goal specification assumptions, and can effectively solve diverse tasks under environmental and reachability constraints.
As opposed to modeling the complex dynamics of the DLO, our method utilizes spatio-temporal information from previous successful experiences, which enables to transfer the trained policy from simulation to the real world. The cooperative control scheme achieves efficient manipulation with keypoint correspondence-based action parametrization. The original contributions of this work are listed as follows:
\begin{itemize}
    \item A contrastive learning-based subgoal planner for long-horizon sparse reward tasks without a goal specification.
    \item A leader-follower control scheme for goal-conditioned collaborative bimanual manipulation under geometric constraints.
    \item A detailed experimental study that evaluates the proposed method in both simulation and real environments.
    % with a real experimental platform.
\end{itemize}
The rest of this paper is organized as follows. Sec. II states the problem formulation. Sec. III introduces the state-action parameterization. Sec. IV explains the policy model. Sec. V reports the results and Sec. VI gives the conclusions. 
% The notation used in this paper is listed in Table. \ref{1-NOTATIONS OF THE PARAMETERS USED IN THIS STUDY}

% \begin{table}[htb]
% \footnotesize
% \centering
% \caption{KEY NOMENCLATURE} % title of Table
% \begin{tabularx}{\columnwidth}{l l} 
% % \hline
% \toprule
% \textbf{Symbol} & \textbf{Description}  \\
% \midrule
% $I^{[t]}$ & the visual observation of the environment $E^{[t]}$ \\
% $Q^{[t]}$ & the detected keypoints of the DLO $L^{[t]}$ \\
% $O^{[t]}$ & the localized obstacles in the observation $I^{[t]}$ \\
% % $L^{[t]},Q^{[t]}$ & the observation and the detected keypoints of the DLO \\
% $S^{[t]},Z^{[t]}$ & the state and the embedding of the environment $E^{[t]}$\\
% $\mathbb{C}_r^v,\mathcal{S}_G$ & the valid configuration of an arm $a_r$ and the goal space \\
% % $\Vec{V}\in\mathcal{V}$ & the action primitive and the space \\
% % $\mathbb{B}$ & the action primitive set \\
% % $\pi(A^{[t]}|S^{[t]},\mathcal{G})$ & the policy model \\
% $\tau\in\mathcal{D}$ & a trajectory in the collected dataset\\
% \bottomrule
% \end{tabularx}
% \label{1-NOTATIONS OF THE PARAMETERS USED IN THIS STUDY} 
% \end{table}

\section{Problem Formulation}
\begin{figure*}
    \centerline{\includegraphics[width=0.9\textwidth]{"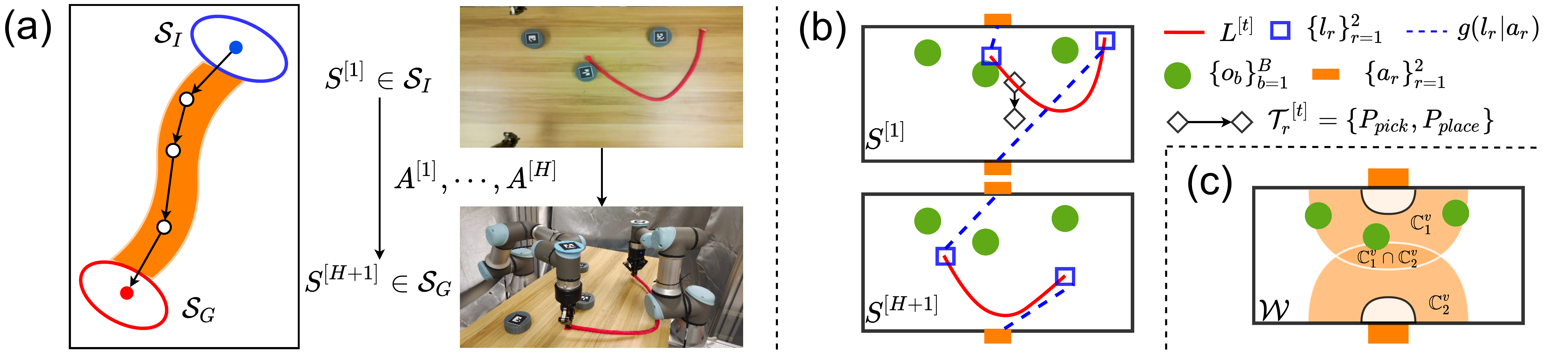"}}
    \caption{
    Schematic diagram of our bimanual manipulation setting.
    (a) Illustration of the context. This task requests robots to rearrange the DLO to enable dual arms to grasp the corresponding ends of the DLO respectively. (b) Graphical representation of the problem formulation. Dual arms are not able to perform prehensile grasping about corresponding ends initially $S^{[1]}$ due to environmental and reachability constraints respectively. Through $H$ pick-and-place sequences $\mathcal{T}_r^{[t]}$, the state of the DLO is transformed as $S^{[H+1]}$ within the goal space $\mathcal{S}_G$.
    % Starting from the initial clutter environment $E^{[1]}$, the purpose of the rearranging task is to shape the DLO to obtain the sparse positive reward $R(S^{[t]},A^{[t]},S^{[t+1]},\mathcal{G})$.  Compared with general goal-conditioned tasks, the sparse positive reward is returned only if the state of the DLO satisfied some conditions (equality and inequality) instead of goal specification.
    (c) Visualization of the valid configuration space $\mathbb{C}_r^{v}\subset\mathbb{C}_r$ of dual arms $\{a_r\}_{r=1}^2$. Each arm $a_r$ only has partial reachability in the complete workspace $\mathcal{W}$, while is also limited by the obstacles. Dual arms share common workspace $\mathbb{C}^{v}_1\cap\mathbb{C}^{v}_2$ in the environment.
    % Due to contact constraints with the obstacles in the clutter environments, the valid configuration space of the arm is constrained as $\mathbb{C}^V_r$. Each arm only has partial reachability in the workspace but share the same space.
    }
    \label{1_overriew}
\end{figure*}
We formulate the problem as a discrete-time 
% finite-horizon discounted 
episodic Markov Decision Process (MDP) represented by a tuple $\mathcal{M} = (\mathcal{S},\mathcal{A},R,\mathcal{P},\rho_0,\gamma,H)$, where $\mathcal{S}$ is the state space, $\mathcal{A}$ is the action space, $R$ is the reward function, $\mathcal{P}(S^{[t+1]}|S^{[t]},A^{[t]})$ is the transition function, $\rho_0$ is the initial state distribution, $\gamma$ is the discount factor and $H$ is the horizon. 
% Instead of goal reaching problem, a specific task is not provided 
Instead of shaping a DLO to a specific configuration defined by a compact descriptor (e.g. a contour \cite{9634846}, a pose \cite{Simeonov2020ALH} or an image \cite{Huo2022KeypointBasedPB}), the objective of this context is to 
% transform an initial state $S^{[1]}$ to 
reach the goal space $\mathcal{S}_G$, a subspace of the state space $\mathcal{S}_G\subseteq\mathcal{S}$ that satisfy geometric conditions.
% to reach a state within the goal space $S^{[H+1]}\in\mathcal{S}_{G}$ (satisfy geometric conditions) from an initial state $S^{[1]}\in\mathcal{S}_{I}$, where the
% through $H$ robot actions, as shown in Fig. \ref{1_overriew}(a).
% requires transforming a state $S^{[1]}\in\mathcal{S}_{I}$ in the initial space to a state inside the goal space $S^{[H+1]}\in\mathcal{S}_{G}$ through $H$ robot actions, as shown in Fig. \ref{1_overriew}(a). 
A typical example in Fig. \ref{1_overriew} is used to introduce the details, which dual arms are not capable of performing prehensile grasping about the corresponding ends of the DLO in the initial state $S^{[1]}$ due to the environmental and reachability constraints. Through $H$ steps of manipulation, the achieved state of the DLO belongs to the goal space $S^{[H+1]}\in\mathcal{S}_G$, which means the geometric conditions are satisfied and the desired grasping is feasible to execute (a real scenario shown in Fig. \ref{1_overriew}(a)).
As shown in Fig. \ref{1_overriew}(b), our unstructured environment $E^{[t]}$ is a rectangle, consisting of a DLO $L^{[t]}$, a set of obstacles $O^{[t]}=\{o_b\}_{b=1}^{B}$, and dual robotic arms $\{a_r\}_{r=1}^2$.
% , defined as $E^{[t]} = L^{[t]}\cup O^{[t]}$, where $L^{[t]}$ is the state of the DLO and $O^{[t]}=\{o_1,\cdots,o_b,\cdots o_B\}$ is the distribution of the obstacles. 
% We define the configuration space of an individual robot $a_r$ as $\mathbb{C}_r$, which can be measured exactly. 
% Note that we assume that the base of the robotic arm is fixed in the environment $E$ and $a_r$ denotes its base position. 
% The valid set of the entire configuration space is denoted as $\mathbb{C}_r^{v}\subset\mathbb{C}_r$, which means the robot is not in collision with the static geometry of the environment $E^{[t]}$ and is reachable according to its physical properties. 
We define the configuration space of an individual robot $a_r$ as $\mathbb{C}_r$ and its corresponding valid set as $\mathbb{C}_r^{v}\subset\mathbb{C}_r$. Validity means that the robot is not in collision with the obstacles in the environment $E^{[t]}$. 
% and is reachable according to its physical properties.
% The problem we want to deal with is to rearrange the configuration of a DLO through planar bimanual manipulation to achieve a binary reward goal. Specifically, we desire the dual arms are capable to grasp the corresponding ends of the DLO under some constraints. In the goal-conditioned setting, the task only provides a sparse extrinsic reward conditioned on a goal $g\in\mathcal{S}$:
% \begin{equation}
%     R(S^{[t]},A^{[t]},S^{[t+1]},g)= \begin{cases}1, & \text { if } S^{[t+1]} == g \\ 0, & \text { else }\end{cases}
% \end{equation}
% For this tasks, we make the following assumptions about the environment $E^{[t]}$:
Following assumptions about the task are made:
\begin{itemize}
    \item The binary mask of the DLO $L^{[t]}$ in the raw visual observation $I^{[t]}$ can be extracted with a color filter.
    \item Each obstacle $o_b$ in the unstructured environment $E$ is static with the same prior size. 
    % with a marker correspondingly.
    \item Both arms $\{a_r\}_{r=1}^2$ only have partial reachability in the complete planar workspace $\mathcal{W}$ but share common region, as shown in Fig. \ref{1_overriew}(c).
    \begin{equation}
    \begin{aligned}
        \mathbb{C}^{v}_1\subset \mathcal{W},\ &\mathbb{C}^{v}_2\subset \mathcal{W}\\
        \mathbb{C}^{v}_1\cap&\mathbb{C}^{v}_2\neq\varnothing
        \end{aligned}
    \end{equation}
    \item There is always a feasible action $A^{[t]}$ for the dual-arm system $\{a_r\}_{r=1}^2$ that deforms the DLO to a new state.
    \begin{equation}
        \exists A^{[t]}\in\mathcal{A},\  S^{[t+1]}=\mathcal{P}(S^{[t]},\ A^{[t]}), S^{[t+1]}\neq S^{[t]}
    \end{equation}
    \item 
    % The mapping $f_G(l_r|a_r)$ between the end of the DLO $l_r$ and the corresponding arm $a_r$ is known in prior. 
    The correspondence between the ends of the DLO $l_r$ and the individual arms $a_r$ is known in prior. 
    % \begin{equation}
    %     l_1 = g(a_1), l_2 = g(a_2)
    % \end{equation}
\end{itemize}

% \emph{2) Action:} 
The action $A^{[t]}$ of bimanual manipulation consists of two sequences with respect to dual arms 
% $A^{[t]}=\{\mathcal{T}^{[t]}_r|\mathcal{T}^{[t]}_1,\mathcal{T}^{[t]}_2\}$
$A^{[t]}=\{\mathcal{T}^{[t]}_r\}_{r=1}^2$, where $\mathcal{T}^{[t]}_r=(P_{pick},P_{place})$ is a pick-and-place sequence and $P=[p_x\ p_y]\in\mathbb{R}^2$ is a two-dimension vector representing the 2D position. Note that we ignore the rotation of the gripper and only consider the straight planar displacement during the whole manipulation, in which the z-value depends on the corresponding pixel in the depth map. 
An action sequence of an arm $\mathcal{T}^{[t]}_r$ is feasible only if all waypoints inside it $P\in\mathcal{T}^{[t]}_r$ are within the corresponding valid configuration $\mathbb{C}^v_r$, denoted as $\forall P\in\mathcal{T}^{[t]}_r: P\in\mathbb{C}_r^v$.
% constrained within its corresponding valid configuration $\mathbb{C}^{v}_r$.
Specifically, whether a planar waypoint $P$ is within the valid configuration $\mathbb{C}^{v}_r$ depends on two inequality conditions:
\begin{equation}
    P\in\mathbb{C}^{v}_r
    \left\{\begin{array}{cc}
\textbf{True} & \epsilon_c < |a_r-P| < \epsilon_f \ \land\\
& \epsilon_o <\min_b |o_b-P|  \\
\textbf{False} & \text { else }
\end{array}\right.
\end{equation}
% \begin{equation}
%     P\in\mathbb{C}^{v}_r
%     \left\{\begin{array}{cc}
% \textbf{True} & \epsilon_c < |a_r-P| < \epsilon_f \epsilon_o <\min_b |o_b-P|  \\
% \textbf{False} & \text { else }
% \end{array}\right.
% \end{equation}
where $|\cdot|$ is the $\mathcal{L}_2$-distance metric, $a_r$ and $o_b$ are the 2D position of $r$-th arm and $b$-th obstacle and $\{\epsilon_c,\epsilon_f,\epsilon_o\}$ are corresponding thresholds of the conditions. 
% According to these constraints, 
It is intuitive that if the minimum and the maximum value of the distance between the waypoints $P\in\mathcal{T}^{[t]}_r$ and the associated instance $\{P_I|a_r,o_b\}$ in the environment match the aforementioned requirements then the sequence $\mathcal{T}^{[t]}_r$ is feasible. The maximum value locates on the edge of the straight movement $D_{max}=\max(|P_I-P_{pick}|,|P_I-P_{place}|)$, while the minimum value $D_{min}$ is computed by:
\begin{equation}
    D_{min}=
    \left\{\begin{array}{cc}
\overrightarrow{P_{1}P_{2}}\times\overrightarrow{P_{1}P_I}/||\overrightarrow{P_{1}P_{2}}|| & \overrightarrow{P_1P_2}\cdot\overrightarrow{P_1P_I}<0 \\ & \land\ 
 \overrightarrow{P_2P_1}\cdot\overrightarrow{P_2P}<0  \\
\min(|P_I-P_{1}|,|P_I-P_{2}|) & \text { else }
\end{array}\right.
\end{equation}
where $P_1$ and $P_2$ represent $P_{pick}$ and $P_{place}$ respectively. 
% To summary, the entire action $A^{[t]} = (P_{pick},P_{place})$ is valid 

% As cooperative bimanual manipulation under unstructured environment, each waypoints $P$ of the sequence $\mathcal{T}^{[t]}$ should belong to the corresponding valid set $\mathbb{C}_r^{v}$. Specifically, two conditions should be satisfied if a waypoint $P$ belongs to the valid set of the individual robot configuration space $\mathbb{C}_r^{v}$, defined by the function $C_r(P,a_r,E^{[t]})$:
% our formulation includes several constraints $C(\pi)$ that restrict the domain of allowed actions $A^{[t]}$ in the environment $E$, including the reachability of the arm $\{a_r|a_1,a_2\}$ and the environmental contacts of the obstacles $O=(o_1,\cdots,o_b,\cdots,o_B)\in\mathcal{O}$

% \emph{3) Policy:} 
The problem we address in this work is how to find a sequence of $H$ feasible robot actions $\{A^{[1]},\cdots,A^{[H]}\}$ under environmental and reachability constraints $(\forall \mathcal{T}_r^{[t]}: \mathcal{T}_r^{[t]}\in\mathbb{C}^{v}_r)$, making the ultimate achieved state of the DLO belongs to the goal space $S^{[H+1]}\in\mathcal{S}_G$. 
As a sparse reward task, the positive signal is returned only when the objective is accomplished, defined as: 
% sparse extrinsic reward $R(S^{[t]},A^{[t]},S^{[t+1]},\mathcal{S}_G)$ of this task is returned only if the resulting environment $S^{[t+1]}$ is within the goal space $S^{[t+1]}\in\mathcal{S}_G$:
\begin{equation}
    R(S^{[t]},A^{[t]},S^{[t+1]},\mathcal{S}_G)= \begin{cases}1, & \text { if } S^{[t+1]}\in\mathcal{S}_G \\ 0, & \text { else }\end{cases}
    \label{reward}
\end{equation}
Specifically, the definition of desired states in the goal space $S^{[t+1]}\in\mathcal{S}_G$ is the endpoints of the DLO located in the valid configuration space of corresponding robotic arms, whose mathematical judgment is:
% about whether a state $S^{[t+1]}$ belongs to the goal space $\mathcal{S}_G$ in this task is:
\begin{equation}
    S^{[t+1]}\in\mathcal{S}_G
    \left\{\begin{array}{cc}
\textbf{True} & l_1\in\mathbb{C}^{v}_1 \land l_2\in\mathbb{C}^{v}_2 \\
\textbf{False} & \text { else }
\end{array}\right.
\end{equation}
% To achieve the goal, we need a policy $\pi(A^{[t]}|E^{[t]})$ that maps from a environment $E^{[t]}$ to an action $A^{[t]}$ to rearrange the a DLO to a new nevironment $E^{[t+1]}$. The objective of the policy $\pi(A^{[t]}|E^{[t]})$ is to properly rearrange the DLO $S^{[t]}$ through the action $A^{[t]}$, making the ends of the DLO $\{s_r|s_1,s_2\},s_r\in S$ are graspable by the corresponding manipulations $\{a_r|a_1,a_2\}$. Specifically, the resulting environment $E^{[t+1]}$ belongs to the goal space $E^{[t+1]}\in\mathcal{G}$ if both $C_1(s^{[t+1]}_1,a_1,E^{[t+1]})$ and $C_2(s^{[t+1]}_2,a_2,E^{[t+1]})$ are satisfied at the same time.
% making the last environment $E^{[H+1]}$ of the episode belonging to the goal space .
% Crucially, the state of the goal space is defined as enabling two arms $\{a_r|a_1,a_2\}$ to grasp the corresponding end of the DLO simultaneously under the constraints of reachability and contact.
% , which maximize the promising reward $R(S^{[t]},A^{[t]},S^{[t+1]},g)$.

% Table. \ref{1-NOTATIONS OF THE PARAMETERS USED IN THIS STUDY} presents the key notation of this paper. 
In summary, the problem we seek to solve is as follows:
\begin{equation}
\begin{aligned}
&\textbf{Find} \ \ \ \ \ \ \ H, A^{[1]},\cdots,A^{[t]},\cdots,A^{[H]} \\
&\textbf{subject to} \ 
% \forall t\in[1,H]: A^{[t]}=\pi(S^{[t]},\mathcal{G}) \\
% & \ \ \ \ \ \ \ \ \ \ \ \ \ 
\forall t\in[1,H]: S^{[t+1]}=\mathcal{P}(S^{[t]},A^{[t]})\\
& 
\ \ \ \ \ \ \ \ \ \ \ \ \ 
\forall t\in[1,H]: \mathcal{T}_r^{[t]}\in\mathbb{C}^{v}_r \\
& \ \ \ \ \ \ \ \ \ \ \ \ \ \ \ \ \ \ \ \ \ 
S^{[H+1]} \in\mathcal{S}_G\\
\end{aligned}
\end{equation}

\section{State-Action Representation}
\subsection{State Parameterization}
One critical issue in vision-based manipulation tasks is how to design an efficient descriptor to extract key features from the visual observation $I^{[t]}$ \cite{8259496}. 
% As a vision-based manipulation task, a critical issue in the context is a low-dimensional descriptor extracting key feature of the environment $E^{[t]}$ from the visual observation $I^{[t]}$ \cite{8259496}. 
% In this article, we leverage semantic keypoints for the state representation due to two reasons: (1) this descriptor is compact and precise; (2) it enables an explanable control scheme in the following policy model.
We use semantic sequential keypoints \cite{Huo2022KeypointBasedPB} for the state representation since they are concise descriptions that allow for an explainable control scheme.
% in the action parameterization. 
% 
% we detect the environment $E^{[t]}$ through visual observation $I^{[t]}$, which is high-dimensional and contains redundant information.
% A precise descriptor about the state $S^{[t]}$ is fundamental for learning the policy $\pi(A^{[t]}|S^{[t]},\mathcal{S}_G)$. As we discussed above, the observation $I^{[t]}$ of the environment $E^{[t]}$ is an image captured by the top-down camera. Learning from observations for this task is challenge owing to the sparse state representation and the high-dimensional DLO. 
% To deal with these issues, we seek to extract the key components of the environment $E^{[t]}$ from the observation $I^{[t]}$.
% adopt pixel coordinates to describe the state $S^{[t]}$. 
% Hence, we seek to extract a compact and efficient state descriptor for the environment, which is fundamental for learning the policy following.
% about the DLO and the unstructured environment is a fundamental condition of the manipulation tasks. 

% Similar to our previous work \cite{Huo2022KeypointBasedPB}, we describe the DLO as a kinematic multi-body system. 
% Different from the mathematical Fourier series-based synthetic dataset in \cite{Huo2022KeypointBasedPB}, we simulate DLOs with the physical engine Bullet \cite{coumans2016pybullet}. We leverage a link-joint structure to describe DLOs and assign the joints as representative keypoints (more details in Sec. V-A). 

Based on the kinematic multi-body model, we describe a DLO with a link-joint structure and designate the joints as representative keypoints. Similar to our previous work \cite{Huo2022KeypointBasedPB}, we detect these sequential keypoints $Q^{[t]}=\{q_k^{[t]}\}_{k=1}^M$ from the mask of the DLO $L^{[t]}$ in the visual observation $I^{[t]}$ through a data-driven neural network $f_D(Q^{[t]}|L^{[t]})$, which is trained on synthetic data. 
% In detail, this network learns to map the mask the DLO $L^{[t]}$ to the sequential keypoints $Q^{[t]}=\{q_k^{[t]}\}_{k=1}^M$ according to the synthetic data.
% maps the labeled keypoints $Q^{[t]}$ to the DLO's mask $L^{[t]}$, where the keypoints are denoted as $Q^{[t]}=\{q_k^{[t]}\}_{k=1}^M$.
% training from synthetic data. 
To simplify the procedures, we render the image-keypoints pair in physical engine Bullet \cite{coumans2016pybullet}, as opposed to the mathematical Fourier series-based model in \cite{Huo2022KeypointBasedPB}. The detailed introduction to the simulation is in Sec. V-A.
In addition, another element that influences the policy $\pi(A^{[t]}|S^{[t]},\mathcal{S}_G)$ is the obstructions $O^{[t]}$ in the environment $E^{[t]}$.
% We express the obstacles $O^{[t]}$ with their location in the observation $I^{[t]}$ in order to share the same representation with the detected keypoints $Q^{[t]}$.
In order to make all the inputs share a common scale without distorting differences in the range of values or losing information, we use the coordinates of the obstacles under observation as their representations.
% it is crucial to encode the information of the obstacles in the environments. Similar to the keypoint representations, each obstacle $o_b$ is described by its coordinate in images.
% about the objects and the environments.
% To reduce the dimension of the environment representation, we use a feature vector instead of the image observation to describe the continuous state, including the keypoint detection $Q^{[t]}$ and the obstacles localization $O^{[t]}$. 
To summarize, we capture the raw image $I^{[t]}$ at each step $t$ and extract the mask of the DLO $L^{[t]}$ with a color filter. Next, we detect the successive keypoints 
$Q^{[t]}=f_D(L^{[t]})$, $Q^{[t]}=\{(q_{ku},q_{kv})\}_{k=1}^M\in\mathbb{R}^{M\times 2}$
and locate the obstacles 
$O^{[t]}=\{(o_{bu},o_{bv})\}_{b=1}^B\in\mathbb{R}^{B\times 2}$
, where $u$ and $v$ are 2D coordinates in image. 
Hence, the representation model $f_R(S^{[t]}|I^{[t]})$ describes the state $S^{[t]}$ of the environment $E^{[t]}$ as $S^{[t]}=(Q^{[t]},O^{[t]})\in\mathbb{R}^{(M+B)\times 2}$.

\begin{figure*}
\centering
    \centerline{\includegraphics[width=0.95\textwidth]{"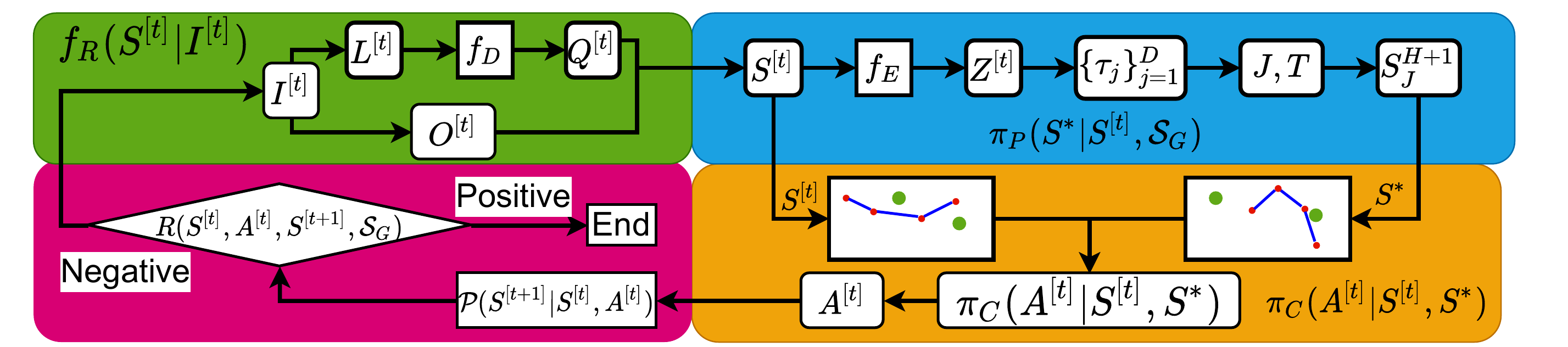"}}
    \caption{Overview of the proposed planning and control framework.
    % for constrained bimanual manipulation without strong assumptions.
    Given the raw observation $I^{[t]}$ at time step $t$, the representation model $f_R(S^{[t]}|I^{[t]})$ extracts the state $S^{[t]}$ of the environment $E^{[t]}$. 
    % Interleaving planning and control, 
    The policy model $\pi(A^{[t]}|S^{[t]},\mathcal{S}_G)$ consists of two modules, global subgoal planning $\pi_P(S^*|S^{[t]},\mathcal{S}_G)$ and local goal-conditioned control $\pi_C(A^{[t]}|S^{[t]},S^*)$. Dual arms execute the action $A^{[t]}$ and the environment transits to $\mathcal{P}(S^{[t+1]}|S^{[t]},A^{[t]})$. The entire policy iterates until the positive reward $R(S^{[t]},A^{[t]},S^{[t+1]},\mathcal{S}_G)$ is obtained.
    }
    \label{2_policy_pipeline}
\end{figure*}

\subsection{Action Parameterization}
% One of the challenges about the policy model $\pi(A^{[t]}|S^{[t]},\mathcal{S}_G)$ is the prohibitively large space of possible actions to accomplish the task, including discrete grasping points and the continuous motion. To remedy the exploration burden, we formulate the pick-and-place sequence of an arm $\mathcal{T}_r^{[t]}=\{P_{pick},P_{place}\}$ as an alignment between the current state $S^{[t]}$ and a desired state $S^*$. 

The unreasonably vast space of potential actions to complete the goal, encompassing discrete grasping points and continuous motion, is one of the challenges with the policy model $\pi(A^{[t]}|S^{[t]},\mathcal{S}_G)$. 
In order to reduce the cost of exploration, we formulate the pick-and-place sequence of an arm $\mathcal{T}_r^{[t]}=\{P_{pick},P_{place}\}$ as correspondence-based manipulation from the present state $S^{[t]}$ to the intended state $S^*$.
% an alignment between the intended state $S^*$ and the presented state $S^{[t]}$.
% current state $S^{[t]}$ and a desired state $S^*$. 
% Specifically, the points of picking and placing are within $P_{pick}\in Q^{[t]}$ and $P_{pick}\in Q^{*}$ respectively.
% Specifically, the picking and placing points are selected within the keypoints of the current state $P_{pick}\leftarrow q^{[t]}_k\in Q^{[t]}$ and the desired state $P_{place}\rightarrow q^{*}_k\in Q^{*}$ respectively.
% The picking and placement locations are specifically chosen inside the keypoints of the intended state $P_{place}\rightarrow q^{*}_k\in Q^{*}$ and the present state $P_{pick}\leftarrow q^{[t]}_k\in Q^{[t]}$, respectively. 
The picking and placement locations are specifically chosen inside the keypoints of the present state $P_{pick}\leftarrow q^{[t]}_k\in Q^{[t]}$ and the intended state $P_{place}\leftarrow q^{*}_k\in Q^{*}$, respectively. 
% within $P_{pick}\in Q^{[t]}$ and $P_{pick}\in Q^{*}$ respectively.
% is one of the keypoints $P_{pick}\in Q^{[t]}$ and the place points is within the keypoints of the goal $P_{pick}\in Q^{*}$. 
% In order to avoid the great change during consective moments, the moving displacement is limited not exceed $\epsilon_D$. Hence, a pick-and-place sequence of an arm is defined as: $\mathcal{T}^{[t]}=(q^{[t]}_k,q^{[t]}_k+\min(\epsilon_D,||\vec{V}||)\cdot\vec{V})$, where the moving direction is $\vec{V} = q^{*}_k-q^{[t]}_k$.
The movement displacement is restricted to not exceed $\epsilon_D$ in order to prevent a significant change during consecutive moments. As a result, the definition of a pick-and-place sequence for an arm is given as follows: $\mathcal{T}_r^{[t]}=(q^{[t]}_k,q^{[t]}_k+\min(\epsilon_D,||\vec{V}||)\cdot\vec{V})$, where the moving direction is $\vec{V} = q^{*}_k-q^{[t]}_k$.

% we implement template-based motion.
% To achieve an efficient, generalizable and flexible bimanual rearranging system, we seek to distangle the action into several representative units, which can be integrated to consist of a complex action. 
% Similar to \cite{PriyaSundaresan2020LearningRM}, the pick-and-place action is formulated with two consecutive moments.

% For the pick-and-place action sequences $A^{[t]}=\{P_{pick},P_{place}\}$
% we narrow down the space of the picking point $P_{pick}$ and the place point $P_{place}$ respectively. The initial pick point $P_{pick}$ is specified as one of the keypoints $q^{[t]}_k\in Q^{[t]}$ of the DLO. The afterward placing point $P_{place}$ is determined by the moving distance $d$ and one of the action primitive $\Vec{V}_i\in\mathcal{V}$ in the displacement set $\mathcal{V}$, defined as:
% \begin{equation}
%     P_{place} = P_{pick} + d\cdot\Vec{V}_i
% \end{equation}
% Specifically, the action primitive space $\mathcal{V}$ includes four alternative instances in pixel space $\mathcal{V}=\{\Vec{V}_i\ |\ [1\ 0],[-1\ 0],[0\ 1],[0\ -1]\}$. Note that we choose the perpendicular directions since they are the most representative in the vector space. 

\section{Policy Model}
In this section, we explain how to learn the policy model $\pi(A^{[t]}|S^{[t]},\mathcal{S}_G)$ to reach the goal space $\mathcal{S}_G$, which maps the current state $S^{[t]}$ to the action $A^{[t]}$.
% conditioned on the goal space $\mathcal{S}_G$.
% A fundamental challenge for the policy $\pi(A^{[t]}|S^{[t]},\mathcal{S}_G)$ is the long-horizon planning complexity without goal specification. As a result, the sample efficiency is very low in this sparse reward setting. 
Due to the long-horizon planning complexity without a goal specification, sampling efficiency is relatively poor in this sparse incentive environment. 
% is a key difficulty for the policy $\pi(A^{[t]}|S^{[t]},\mathcal{S}_G)$. Therefore, , sampling efficiency is relatively poor.
To solve this issue, we factorize the policy $\pi(A^{[t]}|S^{[t]},\mathcal{S}_G)$ into global subgoal planning $\pi_P(S^*|S^{[t]},\mathcal{S}_G)$ and local goal-conditioned control $\pi_C(A^{[t]}|S^{[t]},S^*)$, which planning here is utilized to offer a promising and practicable detailed goal $S^*$ for the local controller to pursue, as shown in Fig. \ref{2_policy_pipeline}. Note that we do not need to design an optimal controller to reach each subgoal exactly; instead, the planner points out the promising direction for the control scheme to move. 
% points out the direction for the controller to approach 
% we only need them to guide the policy training proce
% Be ware that we do not expressly need the policy to reach the goal explicitly; instead, we leverage the goal to direct the policy.
% planning here is formmulated as seeking to switch the state of DLOs to a new neighborhood, making the promising probability that there exist future actions such that the complete sequence leads to achieve the binary goal.
% with respect to the task length.
% As a long-horizon task with sparse rewards, the biggest challenges are agents are not capable of obtaining immediate feedback from
% As a data-driven model, the biggest challenges are the sparse reward of the task, resulting to a low data sample efficiency. Although some references have leverage human demonstrations to deal with this problem, it requires a great amount of time and efforts. In order to deal with this problem without human supervision, we interleave planning and control into our policy model based on a collected dataset in simulation through random actions. In principle, the planner provides a sub-goal for the goal-conditioned local controller. Based on the similarity measurement, the planner determine whether replans or not after executing the action at each step. 
In the following, we first introduce the procedures of data collection in simulation without human participation. Then, we explain how to train the models of subgoal planning based on the collected dataset and the details of the goal-conditioned controller. At last, we explain the pipeline of the policy implementation in real-time after training.

\subsection{Data Collection}
% Since real-world data on robots is expensive, we collect the exploration experiences of robots in simulation. Note that the state $S^{[t]}$ of the environment $E^{[t]}$ is directly accessible in simulation, and the details about the description of the environment and the manipulation of the DLO are described in Sec. V-A.
We collect the exploratory experiences of robots via simulation since real-world data on robots is costly. It should be noted that the simulation provides direct access to the state $S^{[t]}$ of the environment $E^{[t]}$. The details about the description of the environment and the manipulation of the DLO are described in Sec. V-A.
% The data collection is used for training both the goal generation neural network and the local control neural network. 

As a long-horizon task, an entire state-action trajectory of an episode $\tau = ([S^{[1]},A^{[1]},S^{[2]},\cdots,A^{[H]},S^{[H+1]}])$ involves sampling actions and record the observed states to form trajectories. Since our pick-and-place sequence $\mathcal{T}_r^{[t]}=(P_{pick},P_{place})$ is determined based on the current state $S^{[t]}$ and a desired goal $S^{*}$, we record $G$ states within the goal space $\mathcal{S}_G$ to form a dataset $\{S^*_g\in\mathcal{S}_G\}_{g=1}^{G}$ through transforming the DLO with arbitrary actions (including picking points and displacements). 
% without any constraints . 
% and record the achieved state $S^{[t+1]}$ to the dataset once the positive reward $R(S^{[t]},A^{[t]},S^{[t+1]},\mathcal{S}_G)$ is obtained.
% Compared with general goal-conditioned tasks, the sparse extrinsic reward $R(S^{[t]},A^{[t]},S^{[t+1]},\mathcal{S}_G)$ of this task is returned only if the resulting environment $S^{[t+1]}$ is within the goal space $S^{[t+1]}\in\mathcal{S}_G$:
% \begin{equation}
%     R(S^{[t]},A^{[t]},S^{[t+1]},\mathcal{S}_G)= \begin{cases}1, & \text { if } S^{[t+1]}\in\mathcal{S}_G \\ 0, & \text { else }\end{cases}
%     \label{reward}
% \end{equation}
% Each episode $\tau$ is terminated once the positive reward $R(S^{[t]},A^{[t]},S^{[t+1]},\mathcal{S}_G)$ is returned or the exploration steps exceed the allowable horizon $H_{max}$.
 
% As discussed above, a pick-and-place sequence $\mathcal{T}_r^{[t]}=(P_{pick},P_{place})$ is determined based on the current state $S^{[t]}$ and a desired goal $S^{*}$, while we do not have any specific subgoal $S^*$ initially. To deal with this problem, we initialize a dataset $\{S^*_g\in\mathcal{S}_G\}_{g=1}^{G}$ with $G$ goals. Specifically, we manipulate the DLO with arbitrary actions (including picking points and displacements) and record the achieved state $S^{[t+1]}$ to the dataset once the positive reward $R(S^{[t]},A^{[t]},S^{[t+1]},\mathcal{S}_G)$ is obtained.

% Based on the goal dataset $\{S^*_g\in\mathcal{S}_G\}_{g=1}^{G}$, we are capable to implement the alignment action parametization to manipulate the DLO.
In order to avoid time-consuming human supervision, we implement the correspondence-based action randomly based on the goal dataset $\{S^*_g\in\mathcal{S}_G\}_{g=1}^{G}$.
% In order to avoid time-consuming human supervision, we implement the alignment action randomly based on the goal dataset $\{S^*_g\in\mathcal{S}_G\}_{g=1}^{G}$.
% The procedures of an episode include choosing a goal $S^*_g$ from $\{S^*_g\}_{g=1}^{G}$ and sample feasible pairs of keypoints $(q^{[t]}_k,q^*_k)$ to align iteratively.
The procedures of an episode include choosing a goal $S^*_g$ within the dataset $\{S^*_g\}_{g=1}^{G}$ and sample feasible actions from the correspondence-based parameterization to execute iteratively.
% utilize it as the subgoal during the entire episode; (2) Capture the state $S^{[t]}$ at time-step $t$ and sample feasible pairs of keypoints $(q^{[t]}_k,q^*_k)$ under environmental and reachability constraints for the action $A^{[t]}$;
% that are feasible under the constraints; 
% (3) Iterate the policy until the positive reward $R(S^{[t]},A^{[t]},S^{[t+1]},\mathcal{S}_G)$ is obtained or the horizon limit $H_{max}$ is reached. 
% Note that we only store the episode $\tau$ with a positive reward into the dataset $\mathcal{D}$. 
Such arbitrary explorations in planning and control unavoidably generate sub-optimal episodes in the dataset $\mathcal{D}$, resulting to a sub-optimal policy model $\pi(A^{[t]}|S^{[t]},\mathcal{S}_G)$ training on it. To remedy this problem, we explore several times for an episode $\tau_j$ instead of a single trial. Specifically, we explore $\epsilon_P$ subgoals in $\{S^*_g\in\mathcal{S}_G\}_{g=1}^{G}$ and correspondingly implement $\epsilon_C$ times goal-conditioned control for each goal. Among the exploration experiences with $\epsilon_P\times\epsilon_C$ episodes, we save the one with the minimum horizon. Finally, we obtain a dataset $\mathcal{D}$ automatically with $D$ successful episodes $\mathcal{D}=\{\tau_j\}_{j=1}^D$, where an episode is $\tau_j=\{[S_j^{[1]},A_j^{[1]},S_j^{[2]},\cdots,A_j^{[H]},S_j^{[H+1]}]\}$. 

\subsection{Global Subgoal Planning}
% Compared with the general goal-conditioned deformable object manipulation, we are not assess the goal specifications. As a result, the policy model $\pi(A^{[t]}|S^{[t]},\mathcal{G})$ has no clear clues to condition due to the lack of the explicit goal.
% This configuration makes the policy model hard to train due to the lack of the explicit goal. 
% The aim of the subgoal planner $\pi_P(S^*|S^{[t]},\mathcal{S}_G)$ is not to generate a configuration for the policy model to reach explicitly, but rather to move the system towards the goal space $\mathcal{S}_G$ along a promising direction. 
The aim of the subgoal planner $\pi_P(S^*|S^{[t]},\mathcal{S}_G)$ is to point out a promising direction towards the goal space $\mathcal{S}_G$ for the query state $S^{[t]}$ rather than to explicitly produce a configuration to attain.
% Generating a subgoal $S^{[*]}$ from scratch is not practical for three reasons. First, the state of the DLO $S\in\mathcal{S}$ is high-dimensional under some physical constraints. Second, the dynamics of the DLO in unstructured environment is quite complex. Third, the environmental constraints makes the availability of the action between two states difficult to determine. 
However, it is impractical to create a subgoal $S^{[*]}$ from scratch for three reasons. First, the state of the DLO $S\in\mathcal{S}$ is high-dimensional under certain physical restrictions. Second, the dynamics of a DLO in an unstructured context are highly sophisticated. Third, given non-linear environmental and reachability limitations, it is challenging to determine the achievability between an initial state and a goal state.
As a result, we formulate the subgoal planning problem as searching a suitable state from previous exploration $S\in\mathcal{D}$. The benefit of this concept is that we can simply transfer the planner from simulation to reality since there is no need to learn the exact dynamics in this complicated environment.
% there is no need to model the dynamics in this complicated environment, enabling us to transfer the planner from simulation to real directly.
% In order to plan a subgoal without modeling the dynamics of the system, we search a suitable state in the collected dataset $S\in\mathcal{D}$ instead of generating from scratch directly. Hence, the planning model is formulated as searching a suitable state among the collected dataset (thus the executed action during the action is not required). 

% subgoal planning problem as searching 
% In this section, we aim to provide a 
% generate a promising and feasible subgoal for the individual state $S^{[t]}$.
% Generating a goal for a sparse reward task from scratch without any prior knowledge is very challenging. Hence, we formulate the subgoal planning as searching a suitable state in the collected dataset $S^{[t]}\in\mathcal{D}$ instead of generating from scratch. 

The motivation of our search-based subgoal planner is the ultimate accomplished state $S_j^{[H+1]}\in\mathcal{S}_G$ of a successful episode $\tau_j$ is a desirable and feasible goal for the states within this episode $\{S_j^{[t]}\}_{t=1}^{H}\in\tau_j$. Based on this understanding, the obvious strategy for subgoal planning is to find a state in the dataset $S\in\mathcal{D}$ that corresponds to the present state $S^{[t]}$. However, it is challenging to acquire a state in the dataset that is exactly the same as the query in the continuous high-dimensional state space.
% Another issue is that an accurate metric to evaluate the similarity between two states is task-dependent and difficult to design. 
% about how to design the evaluation metric about the similarity between two states.   

Hence, we convert the matching issue during the searching process to clustering within different episodes in the dataset $\{\{S_j^{[t]}\}_{t=1}^{H+1}\}_{j=1}^{D}\in\mathcal{D}$. 
% and design a corresponding task-agnostic similarity metric about the samples in the dataset. 
% However, clustering the states directly is impracticable since the states belong to various episodes are probably share similar features in the original geometric space, making an efficient similarity metric about them is hard to determine.
The states from different episodes are likely to have comparable properties in the original geometric space, making a direct grouping of the states impractical.
To deal with these issues, we utilize a data-driven encoder $f_E(Z^{[t]}|S^{[t]})$ that maps a state $S^{[t]}$ in the geometric space to an embedding $Z^{[t]}$ in the latent space.

We train our encoder $f_E(Z^{[t]}|S^{[t]})$ via a contrastive-learning manner \cite{WilsonYan2020LearningPR}, whose key idea is to make the positive pairs of samples disperse closer while the negative pairs diffuse farther. This setting allows us to re-distribute the samples in the space in accordance with our desired criteria. Also, it is straightforward to employ a universal measure to assess the similarity between samples without task-specific information in the latent space.

\begin{figure}
    \centerline{\includegraphics[width=\columnwidth]{"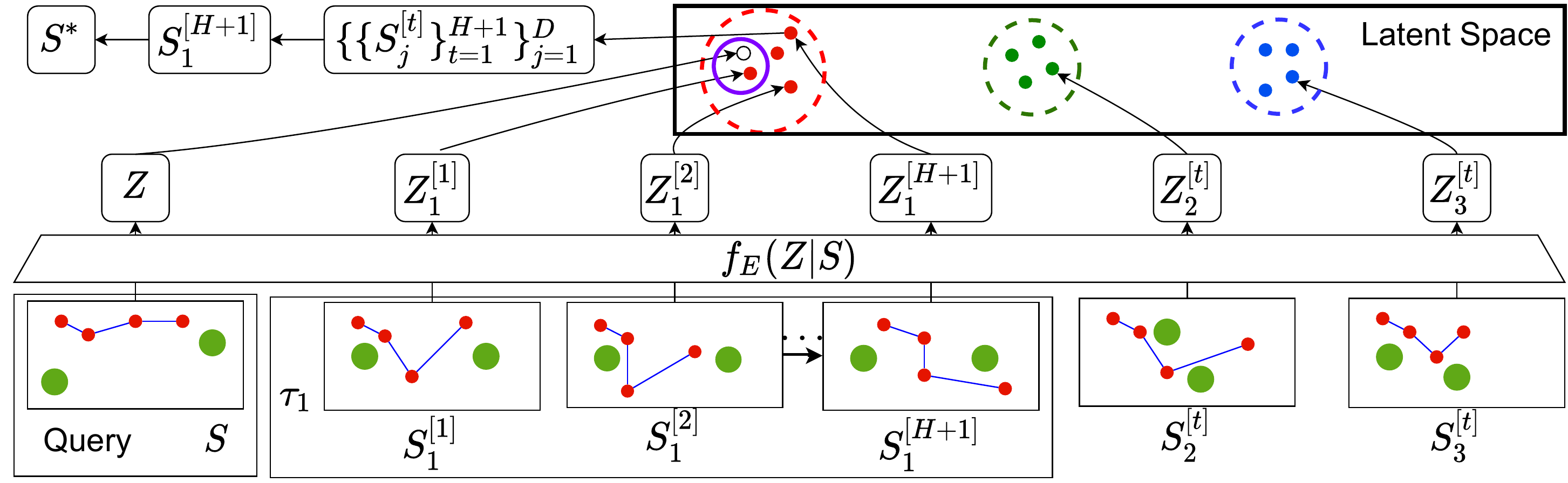"}}
    \caption{Conceptual representation of our contrastive subgoal planning model. Training data of the embedding model consists of $D$ trajectories $\{\{S^{[t]}_j\}_{t=1}^{H+1}\}_{j=1}^{D}$. The objective of the contrastive loss is to bring the positive embedding pairs (belong to the same $j-th$ episode) closer together and the negative embeddings (other states in the dataset $\mathcal{D}$) further away. 
    % The contrastive loss objective brings the positive embedding pairs closer together and the negative embeddings further away.
    }
    \label{3_plan_contra}
\end{figure}
% Fig. \ref{3_plan_contra} illustrates the idea of the training and the prediction of the contrastive learning-based subgoal planner. 
The concept of the training and prediction of the contrastive learning-based subgoal planner is illustrated in Fig. \ref{3_plan_contra}.
For a state in the dataset $S^{[t]}_j\in\{\{S_j^{[t]}\}_{t=1}^{H+1}\}_{j=1}^{D}$, its positive samples are other states belong to the same episode $\{\{S^{[t]}_j\}_{t=1}^{H+1}\setminus S^{[t]}_j \}\in\tau_j$ while its negative samples are other states in the dataset 
% excluding $j-th$ episode 
$\mathcal{D}\setminus \{S_j^{[t]}\}_{t=1}^{H+1}$. 
Notice that the positive and negative associations are not absolute, but rather relative.
% Among them, each state is mapped to the latent space through the encoder $f_E(Z^{[t]}|S^{[t]})$. 
% For each training with respect to a state $S^{[t]}_j\in\mathcal{D}$, we sample a positive state $\overline{S}$ and $N$ negative states $\{\widetilde{S}_n\}_{n=1}^N$ from the corresponding set.
We choose a positive state $\overline{S}$ and $N$ negative states $\{\widetilde{S}_n\}_{n=1}^N$ from the corresponding set for each training with regard to a state $S^{[t]}_j\in\mathcal{D}$.
% a state from other states within the same episode as positive elements $\{\overline{S}^{[t]}_j\in\{S^{[t]}_j\}_{t=1}^{H+1}\setminus \{S_j^{[t]}\}_{t=1}^{H+1}\}$ and sample $N$ states in the remaining dataset as negative elements $\{\widetilde{S}_n\}_{n=1}^N\in\{\mathcal{D}\setminus \{S_j^{[t]}\}_{t=1}^{H+1}\}$. 
With these pairs, we leverage InfoNCE loss \cite{Oord2018RepresentationLW} to train the encoder $f_E(Z^{[t]}|S^{[t]})$:
% \begin{equation}
%     \mathcal{L}_{NCE}=-\mathbb{E}_{\mathcal{D}} \frac{f_H\left(Z^{[t]}_{j},\overline{Z}\right)}{\sum_{n=1}^{N} f_H\left(Z^{[t]}_{j},\widetilde{Z}_n\right)}
%     \label{nce loss}
% \end{equation}
\begin{equation}
    \mathcal{L}_{NCE}=-\mathbb{E}_{\mathcal{D}} \frac{f_H\left(f_E(S^{[t]}_{j}),f_E(\overline{S})\right)}{\sum_{n=1}^{N} f_H\left(f_E(S^{[t]}_{j}),f_E(\widetilde{S}_n)\right)}
    \label{nce loss}
\end{equation}
% where $Z^{[t]}_{j}=f_E(S^{[t]}_{j})$ is the embedding of the state $S^{[t]}$, the positive sample is $\overline{Z}_j^{[t]}=f_E(S^{[t']}_{j}),S^{[t']}_{j}\in\{S_j^{[t]}\}_{t=1}^{H+1}\setminus S_j^{[t]}$, while the negative samples are $\widetilde{Z}_n=f_E(\widetilde{S}_n),\widetilde{S}_n\in\mathcal{D}\setminus \{S^{[t]}_j\}_{t=1}^{H+1}$. 
$N$ is the number of negative samples and $f_H$ is the similarity metric in the embedding space, which we choose bilinear cross product here:
\begin{equation}
    f_H(Z_1,Z_2) = \exp(Z_1^T\cdot Z_2)
    \label{similarity func}
\end{equation}

The motivation behind this learning objective lies in maximizing mutual information between the predicted encodings belonging to the same episode. Within the embedding space, this results in the states belonging to the same episode pairs being placed together but the negative samples pushed further apart. 

% With the embedding encoding $f_E(Z^{[t]}|S^{[t]})$, we maps the states in the dataset $S\in\mathcal{D}$ to the latent space, enabling us to compare the similarity between the states. 

After training the embedding encoding model $f_E(Z^{[t]}|S^{[t]})$, the subgoal planner leverages it to plan a subgoal $S^*$ for the observed state $S^{[t]}$ during the policy implementation. We first process the states of the collected dataset with the encoding models $\{\{Z^{[t]}_j=f_E(S^{[t]}_j)\}_{t=1}^{H+1}\}_{j=1}^{D}$.
To clarify the annotations, a state $S_j^{[t]}$ with $j-th$ subscript comes from the collected dataset $\mathcal{D}$ while a state $S$ is a query during policy implementation.
% the observation during policy implementation.
Fig. \ref{3_plan_contra} illustrates how we obtain a subgoal $S^*$ for a state $S$ in the query during policy execution. 
% prediction process of this subgoal planner during implementation. 
% For a new state $S$ during the manipulation,
At first, we map it to the latent space through the encoder $Z=f_E(S)$. Then, we determine which embedding in the encoded dataset $\{\{Z^{[t]}_j\}_{t=1}^{H+1}\}_{j=1}^{D}$ is most similar to it based on the similarity metric $f_H$:
\begin{equation}
    J, T = \arg\max_{j,t} f_H(Z,Z_j^{[t]})
    \label{get similar one}
\end{equation}
With this equation, we acquire the $T-th$ state of the $J-th$ episode in the dataset $S^{[T]}_J\in\mathcal{D}$ that is most similar to the current state $S$. Hence, we assign the ultimate achieved state $S_{J}^{[H+1]}$ of the $J-th$ episode $\tau_J$ in the dataset as the subgoal for $S$, denoted as:
$S^*\leftarrow S_J^{[H+1]}\leftarrow\pi_P(S^*|S^{[t]},\mathcal{S}_G)$.
% $S^*=\pi_P(S^*|S^{[t]},\mathcal{S}_G)=S_J^{[H+1]}$.

% where $S_j^{[t]}\in\mathcal{D}$. Based on this clustering, we assign the achieved goal $S_J^{[H+1]}$ of $j-th$ episode as the planned goal for the current moment: $S^*=\pi(S^*|S^{[t]},\mathcal{G})=S_J^{[H+1]}$.

% the subgoal planner searches within
% the subgoal planning model output a state $S^*$ for the current state $S^{[t]}$ in an episode based on the similarity metric $f_H$ and the dataset $\mathcal{D}$. Specifically, we first find out which state $S_j^{[t]}$ in the collected positive dataset $\mathcal{D}_P$ whose embedding has the highest similarity with the embedding of the current state $S$: 
% \begin{equation}
%     J, T = \arg\max_{j,t} f_H(f_Z(S),f_Z(S_j^{[t]}))
% \end{equation}
% where $S_j^{[t]}\in\mathcal{D}$. Based on this clustering, we assign the achieved goal $S_J^{[H+1]}$ of $j-th$ episode as the planned goal for the current moment: $S^*=\pi(S^*|S^{[t]},\mathcal{G})=S_J^{[H+1]}$.

\subsection{Local Goal-conditioned Control}
The local controller $\pi_C(A^{[t]}|S^{[t]},S^*)$ is responsible for refining the configuration of the DLO $S^{[t]}$ based on the subgoal $S^*$ supplied by the planner model $\pi_P(S^*|S^{[t]},\mathcal{S}_G)$. Note that we do not expressly need the controller to accomplish the subgoal $S^*$, but rather to manipulate the DLO to approach it in the promising direction. 
% Under this formulation, an action of a arm is defined as: $A^{[t]}=(q^{[t]}_k,q^{[t]}_k+\min(\tau,||\vec{V}||)\cdot\vec{V})$, where the moving direction is $\vec{V} = q^{*}_k-q^{[t]}_k$.

% Compared with general single-arm manipulation in structured environments, our task is much more challenging due to the higher-dimensional discrete-continuous hybrid action spaces for dual arms under environmental and reachability constraints. 
Our task is significantly more difficult than conventional single-arm manipulation in structured contexts due to the higher-dimensional discrete-continuous hybrid action spaces for dual arms under reachability and environmental constraints.
% of reachability and obstacles.
% The biggest challenges of the local controller is the motion planning with respect to dual arms under the systematic constraints.
% There are several key components of the task:
% When determining the action $A^{[t]}$, it is necessary to take into account several issues: 
% Several key components are required to consider when determine the actions $A^{[t]}$:
The following factors must be considered while deciding on the constrained cooperative action $A[t]$ of dual arms:
% (1) The mode switch between a single-arm or bimanual action; (2) The roles assignment between dual arms; (3) The discrete selection of pairs of keypoints; 
% (4) The crash avoision between the motion of dual arms. 
(1) The choice of a single-arm or bimanual action mode; (2) The assignment of duties between dual arms; (3) The discrete choice of keypoint pairings;
(4) The avoidance of collisions when moving two arms simultaneously.
To propose an efficient solution addressing these issues, 
we decouple the roles of twin arms as a leader and a follower \cite{Liu2022RobotCW}.
% criteria to determine the chosen keypoints; 
% , which dual arms in the system are decoupled into different roles (a leader and a follower).

\begin{figure}
    \centerline{\includegraphics[width=\columnwidth]{"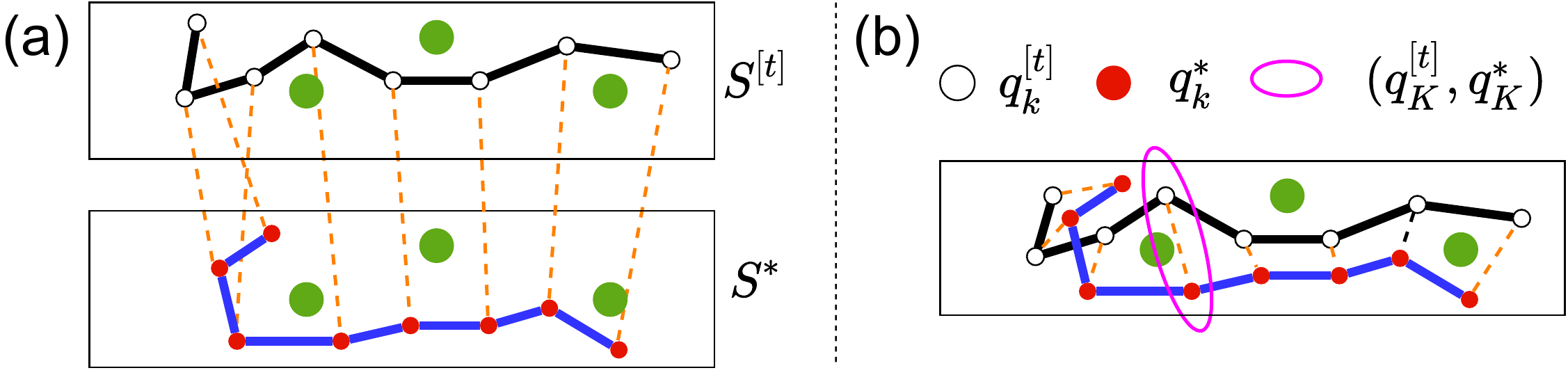"}}
    \caption{Visualization of the leader-follower control scheme. (a) The keypoints of the current step $Q^{[t]}$ and the subgoal $Q^{*}$ are matched one by one. (b) The search for the leader starts from the keypoints pair with largest $\mathcal{L}_2$-norm discrepancy between $Q^{[t]}$ and $Q^{*}$ (shown in purple). 
    % Based on the keypoint correspondence in $S^{[t]}$ and $S^*$, the leader searches around the keypoints pair with largest $\mathcal{L}_2$-norm discrepancy (shown in purple in (b)). 
    }
    \label{4_control}
\end{figure}
% The main differences of our work to \cite{Liu2022RobotCW} are picking points selection at each time-step without pre-grasping and non-fixed roles for each arm.
% In principle, we first find a available keypoint to maximize the deformation as a leader and another keypoint with largest as a follower. 
Both the leader and the follower determine the pick-and-place sequence according to the correspondence of keypoints between $S^{[t]}$ and $S^*$, as shown in Fig. \ref{4_control}(a). Without fixed contacts, dual arms adjust their picking points at each time step and their individual roles can be switched. Note that a follower is not guaranteed to participate in this restricted manipulation
% due to the
% environmental and reachability constraints 
(thus the system switches to single-arm mode). 
Both pick-and-place sequences of them are acquired through optimization subject to constraints, which are discussed in detailed in the following.
% we design a keypoints distribution scheme, whose objective is to maximize the deformation under systematic constraints. 

The objective of the leader is to reduce the error between the current state $S^{[t]}$ and the subgoal $S^*$ as much as possible. Hence, we find out the pair of corresponding keypoints with the highest discrepancy in $\mathcal{L}_2$-distance: $K=\arg\max_k ||q^*_k - q^{[t]}_k||_2$ and the moving direction $\vec{V}$ of the pick-and-place sequence $\mathcal{T}^{[t]}$ is: $\vec{V} = q^*_k - q^{[t]}_k$, shown in Fig. \ref{4_control}(b). 
% However, the pick-and-place sequence $\mathcal{T}^{[t]}$ with respect to the specific $K-th$ pair of keypoints $(q^{[t]}_K,q^{*}_K)$ sometimes is not feasible for any arm $\{a_r\}_{r=1}^2$ due to constraints.
% % of reachability and contacts. 
% Hence, we seek to find a pair of keypoints $(q^{[t]}_{K^*},q^{*}_{K^*})$ that is close to the target pair of keypoints $(q^{[t]}_K,q^{*}_K)$ while an arm is feasible to execute the corresponding pick-and-place sequence.
However, the pick-and-place sequence $\mathcal{T}^{[t]}$ with respect to the specific $K-th$ pair of keypoints $(q^{[t]}_K,q^{*}_K)$ occasionally is not feasible for any arm $\{a_r\}_{r=1}^2$ in the constrained action space. Therefore, we look for a pair of keypoints $(q^{[t]}_{K^*},q^{*}_{K^*})$ that is close to the target pair of keypoints $(q^{[t]}_K,q^{*}_K)$ while still allowing an arm to carry out the corresponding pick-and-place sequence.
% sometimes is not feasible for any arm $\{a_r\}_{r=1}^2$ due to constraints.
% % of reachability and contacts. 
% Hence, we seek to find a pair of keypoints $(q^{[t]}_{K^*},q^{*}_{K^*})$ that is close to the target pair of keypoints $(q^{[t]}_K,q^{*}_K)$ while an arm is feasible to execute the corresponding pick-and-place sequence.
The optimization problem is:
\begin{equation}
\begin{aligned}
&K^*=\arg\min_k ||q^{[t]}_k-q^{[t]}_K||_2 \\
&\textrm{s.t.} \quad \exists a_r\in\{a_r\}_{r=1}^2,  
\forall P\in \mathcal{T}_r^{[t]}, P\in\mathbb{C}^v_r\\
\end{aligned}
\end{equation}
% The solution of this optimization plays the role of a leader. Specifically, 
The chosen arm $a_r$ in this solution acts as the role of a leader, which implement the pick-and-place sequence $\mathcal{T}_r^{[t]}=(q^{[t]}_{K^*},q^{[t]}_{K^*}+\min(\epsilon_D,||\vec{V}||)\cdot\vec{V}), \vec{V} = q^{*}_{K^*}-q^{[t]}_{K^*}$ with respect to the $K^*-th$ keypoint. 

% The objective of the follower is to assist the leader with shaping the DLO globally instead of local area around the selected $K^*-th$ keypoint. 
Designating one arm as the leader, the other arm naturally acts as the follower.
The purpose of the follower is to cooperate with the leader to reshape the DLO as a whole rather than just the surrounding region around the specified $K^*-th$ keypoint.
% Since we have assigned the arm $a_r$ as a leader, another arm $a_{r'}$ is selected as the follower. 
% Since the arm $a_r$ has been designated as the leader, the arm $a r' is chosen as the follower.
% For the pick-and-place sequence $\mathcal{T}_{r'}$, 
We search for $K'-th$ keypoint with the biggest disparity in $\mathcal{L}_2$-distance to the chosen $K^*-th$ keypoint of the leader due to following reasons:
(1) deform the DLO globally ; (2) prevent a collision between the leader and follower.
% Then, another arm $a_{r'}$ acts as the follower. We hope the picking keypoint $q^{[t]}_{k'}$ of the follower far from the leader. The reasons are: 
% (1) Deform the entire DLO globally; 
% % (2)  Manipulate the end more; 
% (2) Avoid the crash between the leader and the follower. 
% Compared with the leader, the moving direction $\vec{V}$ of the follower is defined with respect to the ends around it: $\vec{V} = l_r^* - l^{[t]}_r, r = \arg\min ||l_r-q^{[t]}_k||_2$. 
In contrast to the leader, the follower's movement direction $\vec{V}$ is specified in relation to the endpoints around it: $\vec{V} = l_r^* - l^{[t]}_r, r = \arg\min ||l_r-q^{[t]}_k||_2$.
Taking these into consideration, the optimization problem in terms of the arm acting the role of a follower is:
\begin{equation}
\begin{aligned}
&K'=\arg\max_k ||q^{[t]}_k-q^{[t]}_{K^*}||_2 \\
&\textrm{s.t.} \quad   
\forall P\in \mathcal{T}_r^{[t]}, P\in\mathbb{C}^v_r\\
& \ \ \ \ \ \ \
||q^{[t]}_{K'}-q^{[t]}_{K^*}||_2 > \epsilon
\end{aligned}
\end{equation}
% \begin{equation}
% \begin{aligned}
% K'=\arg&\max_k ||q^{[t]}_k-q^{[t]}_{K^*}||_2 \\
% & \textrm{s.t.} \quad  
% \forall P\in A^{[t]}, P\in\mathbb{C}^v_{r'}\\
% & \ \ \ \ \ \ \ ||q^{[t]}_{K'}-q^{[t]}_{K^*}||_2 > \epsilon
% \end{aligned}
% \end{equation}
where $\epsilon$ is the minimum distance threshold between the picking points of the leader and the follower to avoid the mutual crash. Note that the role of the follower is eliminated if no available solution is obtained (thus switching to single-arm mode).

In summary, the local goal-conditioned controller $\pi_C(A^{[t]}|S^{[t]},S^*)$ outputs the action 
$A^{[t]}=\{\mathcal{T}^{[t]}_r\}_{r=1}^2$ 
to approach the subgoal $S^*$, which dual arms $\{a_r\}_{r=1}^2$ play different roles (leader and follower) for this bimanual manipulation task under environmental and reachability constraints.

\subsection{Policy Implementation}
\begin{algorithm}
	\caption{Policy Implementation}
	\label{Policy Implementation}
	\While{$R(S^{[t]},A^{[t]},S^{[t+1]},\mathcal{S}_G)\neq1$}
	{
	    $S^{[t]}\leftarrow$ $f_R(I^{[t]})$\\
	    $Z^{[t]}\leftarrow$ $f_E(S^{[t]})$\\
	    $J, T \leftarrow \arg\max_{j,t} f_H(Z^{[t]},Z_j^{[t]})$\\
	    $S^{*}\leftarrow$ $\pi_P(S^*|S^{[t]},\mathcal{S}_G)$\\
	    $A^{[t]}\leftarrow$ $\pi_C(A^{[t]}|S^{[t]},S^*)$\\
	    $S^{[t+1]}\leftarrow \mathcal{P}(A^{[t]},S^{[t]})$\\
	}
\end{algorithm}
% In this section, we summarize the policy implementation interleving global subgoal planning and local goal-conditioned control, whose complete pipeline is shown in Alg. \ref{Policy Implementation}.
This section provides an overview of the policy implementation process, which combines global subgoal planning with local goal-conditioned control. The whole policy implementation pipeline is depicted in Alg. \ref{Policy Implementation}.

% At each time-step $t$ in an episode, we start from capturing the image observation $I^{[t]}$ of the unstructured environment $E^{[t]}$. Then, our representation model $f_R(S^{[t]}|I^{[t]})$ extracts the state $S^{[t]}$ through keypoint detection and obstacle localization (Sec. III-A).

We begin by recording the image observation $I^{[t]}$ of the unstructured environment $E^{[t]}$ at each time step $t$ in an episode. Next, our representation model $f_R(S^{[t]}|I^{[t]})$ extracts the state $S^{[t]}$ through keypoint detection and obstacle localization (Sec. III-A).
% extract the representation of keypoints 
% $Q^{[t]}$ and obstacles $O^{[t]}$ to consist of the state $S^{[t]}$ (Sec. III-A) firstly.
Second, the subgoal planner $\pi_P(S^*|S^{[t]},\mathcal{S}_G)$ obtains a promising and feasible state from the collected dataset $S^{[H+1]}_J\in\mathcal{D}$ through similarity matching in the latent space, which is encoded by the contrastive learning-based embedding model $f_E(Z^{[t]}|S^{[t]})$.
% Second, we plan a global subgoal $S^*$ with the planner $\pi_P(S^*|S^{[t]},\mathcal{S}_G)$ based on the encoder $f_E(Z^{[t]}|S^{[t]})$ trained with contrastive learning and the collected dataset $\mathcal{D}$.
Third, taking into consideration the leader and follower roles subject to environmental and reachability constraints, the goal-conditioned controller $\pi_C(A^{[t]}|S^{[t]},S^*)$ determines the cooperative action of dual arms $A^{[t]}=\{\mathcal{T}^{[t]}_r\}_{r=1}^2$.  
% $A^{[t]}=\{\mathcal{T}^{[t]}_r|\mathcal{T}^{[t]}_1,\mathcal{T}^{[t]}_2\}$. 
% of reachability and obstacles.
% we enter the coarse-to-fine action framework (Sec. III-B) to determine the modes and obtain the available action primitives $\mathbb{B}$. 
% \begin{equation}
%     S^* = \arg\max_{S\in\mathcal{D}} f_H(f(S),f(S^{[t]})
% \end{equation} 
% Fourthly, the local controller $\pi(A|S^{[t]},S^*,\mathbb{B})$ output the action probability distribution conditioned on the current state, the planned goal and the feasible set $\mathbb{B}$. Among them, we select the action with the most probability:
% \begin{equation}
%     A^{[t]} = \arg\max_A \pi(A|S^{[t]},S^*,\mathbb{B})
% \end{equation}
At last, dual arms $\{a_r\}_{r=1}^2$ execute the specified action $A^{[t]}$ to transform the state of the environment from $S^{[t]}$ to $S^{[t+1]}$.
% With the selected action, the corresponding agents execute the action in the avaliable set $\mathbb{B}$.
% To adapt the new state after action execution, the whole policy replans the subgoal at each step.
The policy replans the subgoal at each step to adapt to the new situation and iterates as a closed-loop control until the acquired state $S^{[t+1]}=\mathcal{P}(A^{[t]},S^{[t]})$ corresponds to the goal space $S^{[t+1]}\in\mathcal{S}_G$ or the maximum limit of the exploration steps $H_{max}$ is reached. 
% As a closed-loop control, the policy iterates until the achieved state $S^{[t+1]}=\mathcal{P}(A^{[t]},S^{[t]})$ belongs to the goal space $S^{[t+1]}\in\mathcal{S}_G$ or the maximum limit of the exploration steps $H_{max}$ is reached. 

\section{Results}
\begin{figure}
\begin{floatrow}
% \hspace*{-0.5cm}
\ffigbox[0.5\columnwidth]
{%
  \includegraphics[width=\columnwidth]{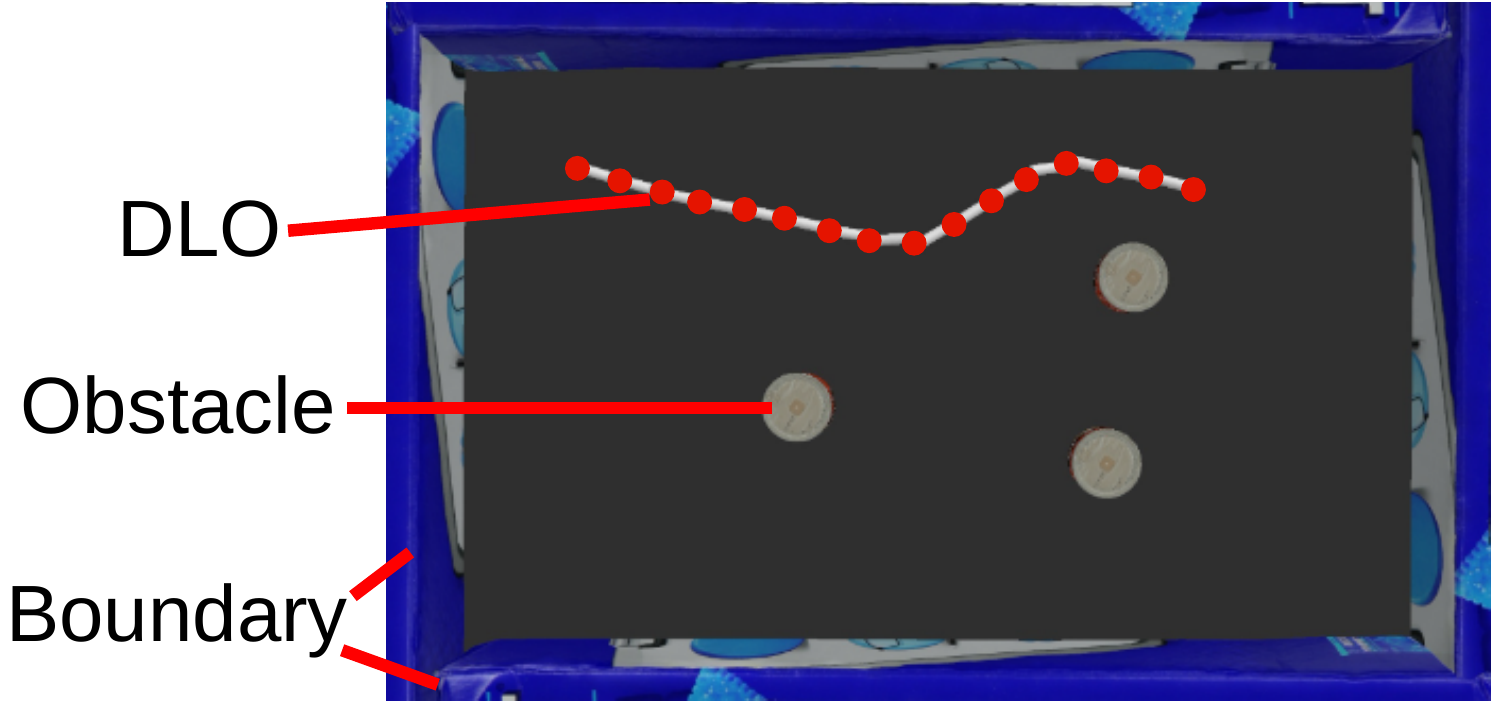}
%   \rule{3cm}{3cm}%
}{%
  \caption{ A snapshot of the simulation.}
  \label{5_simulation_env}
}
% \hspace*{0.5cm}
\capbtabbox{%
% \hspace*{1cm}
\footnotesize
  \begin{tabular}{cc}\hline
  Paras & Spec. \\ \hline
    % length of DLO & [0.5,0.7] \\
    $\lambda$ & $[0.5m,0.7m]$ \\
    $M$ & $16$ \\
    % joint type & revolute \\
    $\epsilon_l$ & $1\ \text{rad}$ \\
    % Obstacle radius & 0.04 \\
    $\mu$ & $0.04$ \\
    % distance between two arms & $0.75m$ \\
    % gripper size & 0.05 \\
    $\epsilon_c,\epsilon_f$ & $0.15m,0.45m$ \\
    $\epsilon_o$ & $0.1m$ \\
    \hline
  \end{tabular}
}{%
  \caption{Task parameters.}
  \label{table_simulation_paras}
}
\end{floatrow}
\end{figure}
% \begin{figure*}
% \begin{floatrow}
% \ffigbox{%
%   \includegraphics[width=\columnwidth]{"figures/5_simulation_env"}
% %   \rule{3cm}{3cm}%
% }{%
%   \caption{ A snapshot of the simulation in Pybullet.}
%   \label{5_simulation_env}
% }
% \capbtabbox{%
% %   \footnotesize
%   \begin{tabular}{cc}\hline
%   Parameters & Specification \\ \hline
%     length of DLO & 0.5m - 0.7m \\
%     joint number & 16 \\
%     % joint type & revolute \\
%     joint angle limit & $[-1\ rad,1\ rad]$ \\
%     Obstacle number & 4 \\
%     Obstacle radius & 0.04m \\
%     workspace size & $1m\times0.7m$ \\
%     % distance between two arms & $0.75m$ \\
%     gripper size & 0.05m \\
%     reachability constraint & [0.15m,0.45m] \\
%     obstacle constraint & 0.1m \\
%     \hline
%   \end{tabular}
% }{%
%   \caption{Physical parameters in simulation.}
%   \label{table_simulation_paras}
% }
% \end{floatrow}
% \end{figure*}

% In this section, we evaluate our method both in simulation and in real worlds. We first introduce the settings of simulation about deformable object manipulation. Then we evaluate the overall performance of our proposed framework interleaving perception, planning and control followed by the second and the third experiments performing statistical comparisons against several ablations and baselines. Finally, we present the results of our algorithm in real worlds without any simulation-to-real fine tuning.

% To evaluate our approach, we perform statistical comparisons of our method versus ablations and baselines in simulation over the context of constrained bimanual manipulation. 
In order to assess our strategy, we undertake statistical comparisons of our technique against baselines and ablations in the setting of constrained bimanual manipulation.
We begin by outlining the simulation conditions and an analysis of our framework that interleaves planning and control using an ablation study. 
% We first present the specification of the simulation settings and an ablation study of our interleaving planning and control framework. 
Then we compare our proposed complete framework with several baselines.
% introduce several chosen baselines and compare our proposed framework with them. 
Finally, we demonstrate the performance of our approach in real-world situations without any fine-tuning.

\subsection{Simulation Setting}
% Our simulation environment is based on Deformable Ravens \cite{seita_bags_2021} using Pybullet \cite{coumans2021}. 
% Simulation has been widely used in data-driven robotics, while only few of them includes deformable objects due to the difficult modeling. 
Data-driven robotics has utilized simulation extensively, but few of them include deformable objects. Through Pybullet \cite{coumans2016pybullet}, \cite{seita_bags_2021} represents a DLO as a sequence of rigid bodies subject to the limitations of a fixed distance between adjacent bodies. 
% Although this representation simulate the articular structure of DLOs, it do not consider the angular damping and the elasicity of them.
However, such an articular structure neglects the angular damping and elasticity of DLOs, deviating its dynamics greatly from actual situations.
% the real cases. 
% To simulate the elasticity of the rope, we set the angle limits of the joints to simulate the stable situation of the rope after shaping. 
In our setting, we simulate a DLO with a kinematic link-joint model that the rotation range of a joint is limited within $[-\epsilon_l,\epsilon_l]$.
% based on the kinematic link-chain model with the Unified Robotics Description Format (URDF). 
% In order to make the simulated object more similar to the elastic rope, we apply a link-chain structure for the cable. 
Specifically, the model comprises $M-1$ equivalent-length links connecting with $M-2$ revolute joints, plus two additional virtual joints to represent the endpoints.
% To render the endpoints of a DLO, we add two virtual joints.
% The endpoints of a DLO are represented by two virtual joints
% In order to connect the initial link and the last link—which stand in for the ends of DLOs—through two joints, two virtual connections are created.
% Two virtual links are added to connect with the first link and the last link through two joints, representing the ends of DLOs. 
Visualized in Fig. \ref{5_simulation_env}, a DLO with a length of $\lambda$ and multiple cans with a radius of $\mu$ serving as obstacles are included in the workspace with a size of $0.6m\times1m$.
% in a locates in the workspace with a size of $0.6m\times1m$ and the length of the DLO $\lambda$ varies from $0.5m$ to $0.7m$. 
% In addition, multiple cans with a radius of $\mu$ serve as obstacles in the clutter environment. Fig. \ref{5_simulation_env} visualizes the simulation and 
Table \ref{table_simulation_paras} lists the specification of the physical parameters.

% The link set includes two virtual links to connect to the joints simulating the end of the DLO.
% , which is corresponding to a keypoint in our representation model respectively.  
% cylinders, which the adjacent pair of links are connecting by a revolute joint. To simulate the elasticity of the rope, we set the angle limits of the joints to simulate the stable situation of the rope after shaping. 

A suction gripper carried by a UR5 robotic manipulator is employed in \cite{seita_bags_2021} to manipulate through virtual attachment restrictions. 
% constraints in simulation. 
% In detail, they attach the end-effector with the DLO through some constraints between the end-effector manipulate.
This approach has two shortcomings:
% There are two drawbacks of this methods: 
(1) The performance of such restrictions is far from the execution of prehensile grasping in actual scenarios. (2) The efficiency of data collecting is decreased by motion planning of robotic arm. 
% To address this problem, 
With the link-joint structure, we simulate the multi-body dynamics  through applying virtual force, whose configuration characterized by the joints $\mathbf{q}\in\mathcal{R}^M$ is modeled as \cite{Chang2020Sim2Real2SimBT}: 
% \cite{NavarroAlarcon2014EnergySM}:
% \begin{equation}
%     \mathbf{M}(\mathbf{p}) \mathbf{p}+\mathbf{C}(\mathbf{p}, \mathbf{p}) \mathbf{p}=\tau_{g}(\mathbf{p})+\mathbf{B u}
% \end{equation}
% \begin{equation}
%     \mathbf{M}(\mathbf{q}) \ddot{\mathbf{q}}+\mathbf{c}(\mathbf{q}, \dot{\mathbf{q}})+\boldsymbol{g}(\mathbf{q})=\boldsymbol{\tau}+\mathbf{J}(\mathbf{q})^{\top} \boldsymbol{f}
% \end{equation}
\begin{equation}
    M \ddot{\mathbf{q}}+C \dot{\mathbf{q}}+G+\mathbf{J}^{T} \mathbf{f}_{e x t}+\mathbf{K} \mathbf{q}+\mathbf{D} \dot{\mathbf{q}}=\boldsymbol{\omega}
\end{equation}
where $\mathbf{q},\dot{\mathbf{q}},\ddot{\mathbf{q}}$ represent joint position, velocity and acceleration. $M$ is the inertial matrix, $\mathbf{C}$ is the centrifugal and Coriolis forces matrix, and $G$ is gravitational forces. $\mathbf{J}^T$ is the transpose of the robot Jacobian and $f_{ext}$ is the external force. $\mathbf{K}$ and $\mathbf{D}$ are the stiffness and damping of the DLO respectively. $\omega$ is a vector of joint torques. 
% $\boldsymbol{g}(\mathbf{q})$ is the gravity vector, $\boldsymbol{\tau}$ are the joint actuation torques and $\boldsymbol{f}$ denotes the external wrench through the Jacobian matrix $\mathbf{J}(\mathbf{q})$.
% . The matrix $\mathbf{C}$ maps control inputs $\mathbf{u}$ into generalized forces. 
% Hence, we apply virtual forces on DLOs without arms in the environment to simulate pick-and-place manipulation by 2-fingered grippers.
In manipulation with our proposed leader-follower control scheme, the external force is parameterized by its position and magnitude, which is maintained until the intended displacement distance $||\vec{V}||$ is reached. 
% Although there is still a gap between the simulation and real environments, this simulation provides effective data to train our policy, allowing us to transfer from simulation to real environments without any fine-tuning.
The designed simulation environment provides effective data for our data-driven model, allowing us to transfer from simulation to real environments without any fine-tuning.
% \subsection{Data Collection}
% We collect data in our experiment.

% \blindtext

\subsection{Ablation Study}
\begin{figure*}
\centering
    \centerline{\includegraphics[width=0.95\textwidth]{"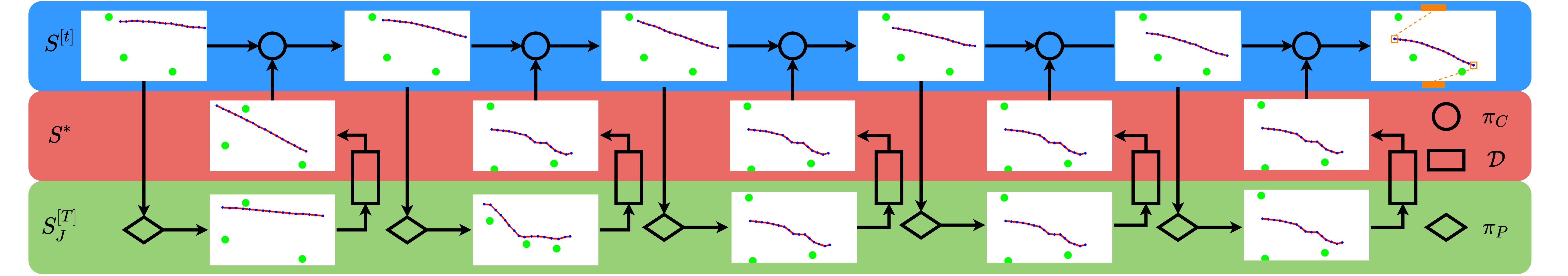"}}
    \caption{A complete trajectory for an episode within the simulator. At each time step $t$, the subgoal planning model $\pi_P(S^*|S^{[t]},\mathcal{S}_G)$ finds the most similar state $S^{[T]}_J$ in the dataset $\{\{S^{[t]}_j\}_{t=1}^{H+1}\}_{j=1}^{D}$ 
    % $\{S_j^{[t]}\}_{t=1}^{H+1}\in\tau_j$ 
    (green row) concerning the current state $S^{[t]}$ (blue row) and assign the corresponding achieved goal of the $J-th$ episode $S_J^{[H+1]}$ as a subgoal $S^*$ (red row). Then, the controller $\pi_C(A^{[t]}|S^{[t]},S^*)$ determines the action $A^{[t]}$ conditioning on the current state $S^{[t]}$ and the subgoal $S^*$. 
    }
    \label{6_mani_process}
\end{figure*}

\begin{figure}
\centering
    \centerline{\includegraphics[width=\columnwidth]{"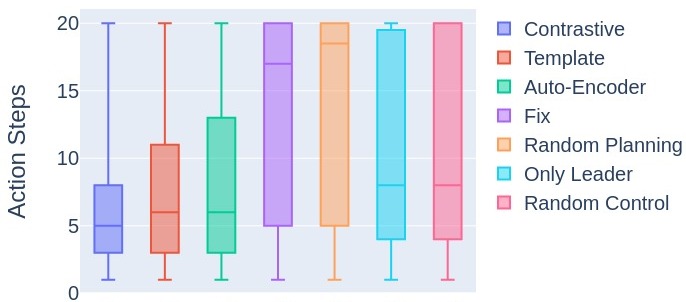"}}
    \caption{Results of comparisons in the ablation study. The boxplot pictures the required action steps of bimanual manipulation. The central straight line corresponds to the median value of the steps, whereas the sides of the box refer to the first and the third quartiles of the data.
    }
    \label{7_box_plot}
\end{figure}

In this experiment, we first exhibit the manipulation procedure of an episode employing our complete framework, which consists of global subgoal planning $\pi_P(S^{*}|S^{[t]},\mathcal{S}_G)$ and local goal-conditioned control $\pi_C(A^{[t]}|S^{[t]},S^{*})$. 
% Then, we implement an ablation study versus individual modules to highlight their and necessity and superiority.
Then, in order to emphasize their excellence and necessity, we conduct ablation research versus individual modules.
% our proposed framework, consisting of global planning $\pi_P(S^{*}|S^{[t]},\mathcal{S}_G)$ and local control $\pi_C(S^{[t]}|S^{[t]},S^{*})$.
% evaluate the performance of the complete framework, consisting of representation $f_R(S^{[t]}|I^{[t]})$, planning $\pi_P(S^{*}|S^{[t]},\mathcal{S}_G)$ and control $\pi_C(S^{[t]}|S^{[t]},S^{*})$. 

A complete episode of our constrained bimanual manipulation is shown in Fig. \ref{6_mani_process}. 
% illustrates a complete episode of our constrained bimanual manipulation task. 
% At each time-step $t$, we represent the environment $E^{[t]}$ as a state $S^{[t]}$ based on the visual observation $I^{[t]}$. 
We describe the environment $E^{[t]}$ as a state $S^{[t]}$ based on the visual observation $I^{[t]}$ at each time step $t$.
Next, we retrieve the embedding $Z^{[t]}$ of the state $S^{[t]}$ with the encoder $f_E(Z^{[t]}|S^{[t]})$ and then locate the most comparable embedding $S^{[T]}_J$ in the dataset $\mathcal{D}$ with Eq. \ref{get similar one}. Then, we assign the achieved goal $S_J^{[H+1]}$ of the $J-th$ episode in the dataset as a subgoal $S_J^{[H+1]}\rightarrow S^*$. At last, the local goal-conditioned controller $\pi_C(A^{[t]}|S^{[t]},S^*)$ takes the current state $S^{[t]}$ and the planned subgoal $S^*$ as input and output the correspondence-based action $A^{[t]}$. The entire planning and control framework iterates until the attached state inside the goal space $S^{[t+1]}\in\mathcal{S}_G$. Following, we put two ablation case studies into practice concerning the planning and the controller, respectively.
 
% \begin{figure}
% \centering
%     \centerline{\includegraphics[width=\columnwidth]{"figures/ablation_box"}}
%     \caption{The procedure of a typical example of our rearranging DLOs with bimanual manipulation. At each time-step, the policy replans the subgoal for the goal-conditoned controller.
%     }
%     \label{ablation}
% \end{figure}

\begin{figure}
\centering
    \centerline{\includegraphics[width=\columnwidth]{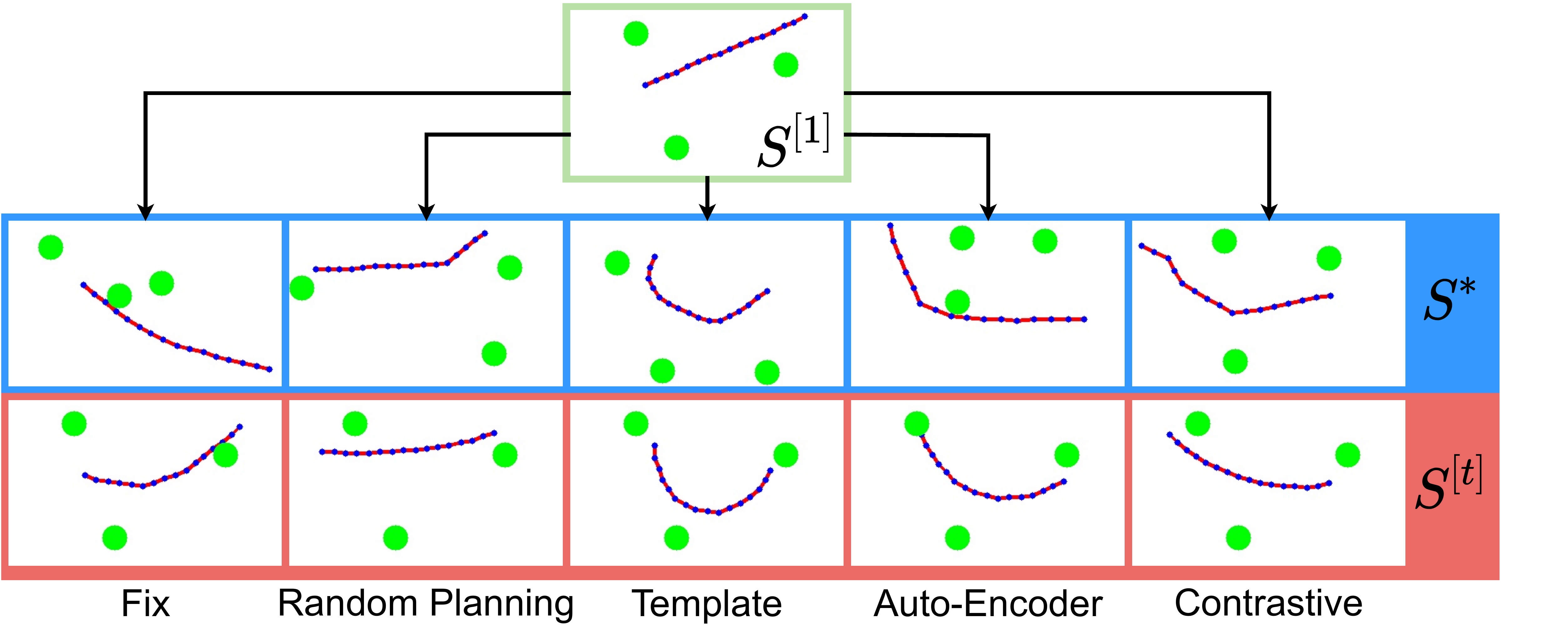}}
    \caption{Examples for each of the ablation studies regarding the subgoal selection. All starting from the same initialization state $S^{[1]}$ (green), different algorithms reach different state $S^{[t]}$ (red) owing to various subgoals $S^{*}$ (blue).
    % , different algorithms reaches to different states (red). 
    % The procedure of a typical example of our rearranging DLOs with bimanual manipulation. At each time-step, the policy replans the subgoal for the goal-conditoned controller.
    }
    \label{7_ablation_plan}
\end{figure}

%tiny scriptsize footnotesize small
\begin{table}
\footnotesize
\centering
% \caption{Comparison Results with various baselines under the planning and control framework }
\caption{Quantitative comparisons between different methods in the ablation study regarding planning and control}
\ADLnullwidehline
\arrayrulecolor{black}
\begin{center}
\begin{tabular}{c|c|ccc} 
\hline
\textbf{Planning} & \textbf{Control} & Success & Mean & Std  \\ 
 & & Rate \% & actions  & actions  \\ 
\hline
\textbf{Fix} & \textbf{LF} & 51.3 & 12.8 &  7.6  \\
\textbf{RP} & \textbf{LF} & 50.5 & 12.8 & 7.7  \\
\textbf{Template} & \textbf{LF} & 72.5 & 9.7 & 7.2  \\ 
\textbf{Auto-Encoder} & \textbf{LF} & 69.8 & 10.2 & 7.3 \\ 
\arrayrulecolor{black}\hdashline
\textbf{Contrastive} & \textbf{OL} & 78.4 & 10.1 & 6.5  \\ 
\textbf{Contrastive} & \textbf{RC} & 79.0 & 10.1 & 6.5 \\
\arrayrulecolor{black}\hdashline
\textbf{Contrastive} & \textbf{LF} & \textbf{86.6} & \textbf{8.0} & \textbf{6.0} \\
\hline
\end{tabular}
\end{center}
\begin{tablenotes}
\item 
\footnotesize{
\textbf{RP}: Random Planning;
\textbf{LF}: Leader-Follower; \textbf{OL}: Only Leader; \textbf{RC}: Random Control.}
\end{tablenotes}
\arrayrulecolor{black}
\label{table_ablation_study}
\end{table}
% \footnotesize{
% \textbf{LF}: Leader-Follower; \textbf{OL}: Only Leader; \textbf{RC}: Random Control.}

First, we claim that our contrastive learning-based planner is capable of extracting the spatio-temporal information of a successful episode, thus providing an appropriate goal for the controller to explore. To back up this assertion, we contrast four alternative subgoal planners, including: (1) \textbf{Fixed}: Pick a fixed subgoal randomly for all episodes; (2) \textbf{Random Planning}: Sample a subgoal randomly for each episode; (3) \textbf{Template}: Choose a subgoal based on $\mathcal{L}_2$-distance metric in geometric space; (4) \textbf{Auto-Encoder}: Obtain a subgoal based on $\mathcal{L}_2$-distance metric in latent space, whose encoder is trained with a self-reconstruction loss. Note that all these contrast algorithms select a subgoal inside the same collected dataset $\mathcal{D}$ and our proposed leader-follower controller is followed to finish the task.
% Note that all the baselines learn and select the goal according to the same collected dataset. 

% Plan the goal with a reconstruction-based auto-encoder. 
% Our proposed leader-follower controller is used to finish the task with respect to any planning method.
% The reasons for these comparisons are to illustrate that a suitable subgoal is significant for the goal-conditioned controller to finish the task.
These comparisons are being made to indicate how important a proper subgoal is for the goal-conditioned controller to complete the assignment.
% Table \ref{table_ablation_study} shows the quantitative results in the comparison and Fig. \ref{7_box_plot} compares in the boxplot. 
The quantitative results of the comparison are shown in Table \ref{table_ablation_study} and the boxplot comparison is shown in Fig. \ref{7_box_plot}.
% These findings reveal that our suggested contrastive approaches achieves the highest success rate among them while requiring fewest amount of necessary action steps. To analyze the results, we visualize a typical example in the comparisons, shown in Fig. \ref{7_ablation_plan}. 
These results show that our suggested contrastive learning-based approaches get the best success rate among them while demanding the least number of required action steps. To analyze the results, we depict a typical case in the comparisons, as shown in Fig. \ref{7_ablation_plan}. This example illustrates that the subgoal provided by our contrastive planner is acceptable and points out a promising direction to reach the goal space for the presented state to approach, while other methods are not practical.
% visualizes a typical example in the comparisons. 
Without any preference, both \textbf{Fix} and \textbf{Random Planning} are unable to offer effective and promising subgoals for the controller.
\textbf{Template} and \textbf{AutoEncoder} operate admirably in some circumstances (achieving a success rate of roughly $70\%$ overall), but struggle when the DLO is close to the barriers.
% \textbf{Template} and \textbf{AutoEncoder} do well in some scenarios (achieved about $70\%$ success rate in total), but perform poorly when the DLO is close to the obstacles. 
% obtains goals with high similarity in geometric and latent space respectively.
% This is mainly likely because the dynamics of the system is not continuous due to the contacts between the DLO and the fixed obstacles. 
The dynamics of the DLO is not smooth as a result of the interactions between it and the fixed obstacles, which makes this mostly plausible.
% The reasons of the low performance is probably due to the existence of obstacles, making the transitions of the system is not continuous. 
Through clustering the states within an episode together, our contrastive learning-based subgoal planner learns the spatio-temporal representation in the success experiences, namely the feasible and promising transition towards the goal space under constraints. Hence, the intended subgoal is better suited to the query state to approach the goal space $\mathcal{S}_G$.
% Through training the embedding model in a contrastive manner, the states distribute closer in the latent space due to the shared spatial-temporal transition in a successful episode, meaning that the corresponding achieved state is achievable and promising for the states similar to this cluster. 
% within a successful episode sharing the same temporal-spatial information distribute closer in the latent space, enabling a better matching. 

\begin{figure}
\centering
    \centerline{\includegraphics[width=\columnwidth]{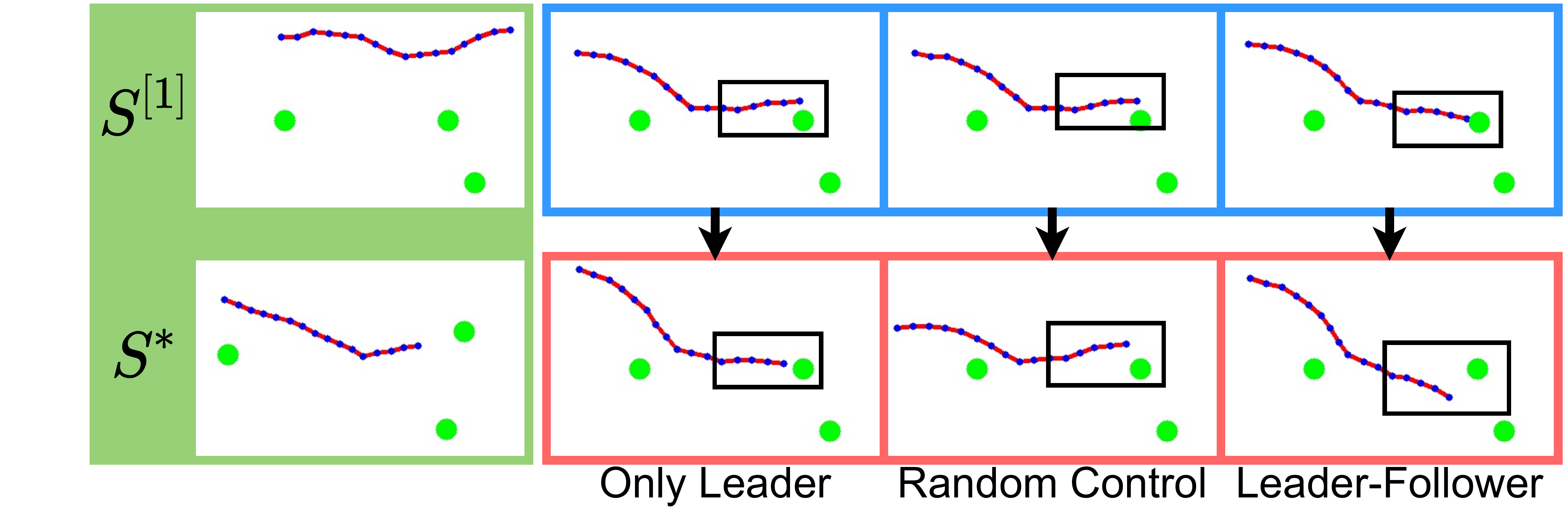}}
    \caption{Examples for each of the ablation studies regarding the goal-conditioned controller. All shares the same initialization state $S^{[1]}$ and the subgoals $S^*$ (green). Our proposed leader-follower scheme is capable to manipulate the DLO globally to move far from the obstacle (black box), while other baselines perform worse.
    }
    \label{8_ablation_control}
\end{figure}

Second, we argue that our leader-follower control scheme enhances the goal-reaching capacities of the DLO under environmental and reachability constraints by utilizing the cooperative skills of bimanual manipulation.
% Second, we argue that our leader-follower scheme takes advantage of the cooperation capabilities of bimanual manipulation $\{a_r\}_{r=1}^2$ to improve the goal-reaching performance of the DLO under environmental constraints. 
% is capable of arranging the action of dual arms $\{a_r\}_{r=1}^2$ to improve the control performance under systematic constraints. 
To substantiate this assertion, we contrast two different controllers, including: 
(1) \textbf{Random Control}: Sample a corresponding-based $(S^{[t]},S^*)$ action $A^{[t]}$ randomly for dual arms $\{a_r\}_{r=1}^2$. 
(2) \textbf{Only Leader}: Only the leader action in our leader-follower framework is executed (single-arm policy). 
Note that all of the alternative controllers execute the action with the same distance threshold $\epsilon_D$ under constraints and obtain subgoals depending on our proposed contrastive learning-based planner.
% Note that our suggested contrastive learning-based planner is employed to generate the subgoal, and that they all carry out the action with the same distance threshold $epsilon D$ under constraints.
% We only use the leader in our leader-follower framework to act as a single-arm action; 
% (2) \textbf{Random Control}: 
% Sample several pick-and-place sequences $\mathcal{T}_r^{[t]}$ for dual arms $\{a_r\}_{r=1}^2$ under systematic constraints.
% Note that all of them execute the action with the same distance threshold $\epsilon_D$ and our proposed contrastive learning-based planner is used to provide the subgoal.

% keypoints $\{q_k^{[t]}\}_{k=1}^M$ for dual arms randomly under constraints. 
% The reasons for this comparisons are to illustrate the efficiency of our leader-follower scheme in terms of goal-conditioned control. Table \ref{table_ablation_study} shows the quantitative results in the comparison and Fig. \ref{7_box_plot} compares in the boxplot. 
The comparisons are made in order to show how effective our leader-follower system is in achieving goal-conditioned control. The quantitative results of the comparison are shown in Table \ref{table_ablation_study} and the boxplot comparison is shown in Fig. \ref{7_box_plot}.
These findings show that among them, our suggested leader-follower scheme has the best success rate while requiring the fewest amount of necessary action steps. A typical example in the comparison is shown in  Fig. \ref{8_ablation_control}, where each baseline has the identical initial state $S^{[1]}$ and subgoal $S^*$. 
% visualizes a typical example in the comparisons, which all of the baselines share the same initialization state $S^{[1]}$ and conditioned subgoal $S^*$. 
% The key challenge of this episode is to move the right end of the DLO away from the obstacle (shown in black box).
Moving the right end of the DLO away from the obstruction is the main barrier of the episode (shown in black box).
% \textbf{Only Leader} only manipulates the right part of the DLO, while the global shape of the object is not manipulated to bypass .
% Owing to the arbitrary interests, \textbf{Random Control} do not have the capability of locating the key parts to manipulate. 
\textbf{Random Control} is unable to find the essential components to control because of the arbitrary interests.
% choose pairs of keypoints randomly to align, making it lose to locate the key position of the object to manipulate. 
% requires additional steps to
% Although \textbf{Only Leader} is still able to manipulate the key part, it can not deform the DLO globally, thus requiring additional steps to approach the subgoal. 
Even though the fact that \textbf{Only Leader} can still alter the key component, it is unable to deform the DLO globally, necessitating extra steps to approach the subgoal.
% only manipulates the right part of the DLO, while the global shape of the object is not manipulated to bypass .
On the contrary, our proposed leader-follower scheme deforms the configuration of the DLO globally with the cooperation between dual arms. 
% Not only does this method reduce the risk of collision between dual arms, but also deforms the DLO towards the subgoal efficiently.
This technique effectively alters the DLO toward the subgoal while simultaneously lowering the chance of collision between dual arms.

\subsection{Comparisons to baselines}
\begin{figure}
\centering
    \centerline{\includegraphics[width=0.9\columnwidth]{"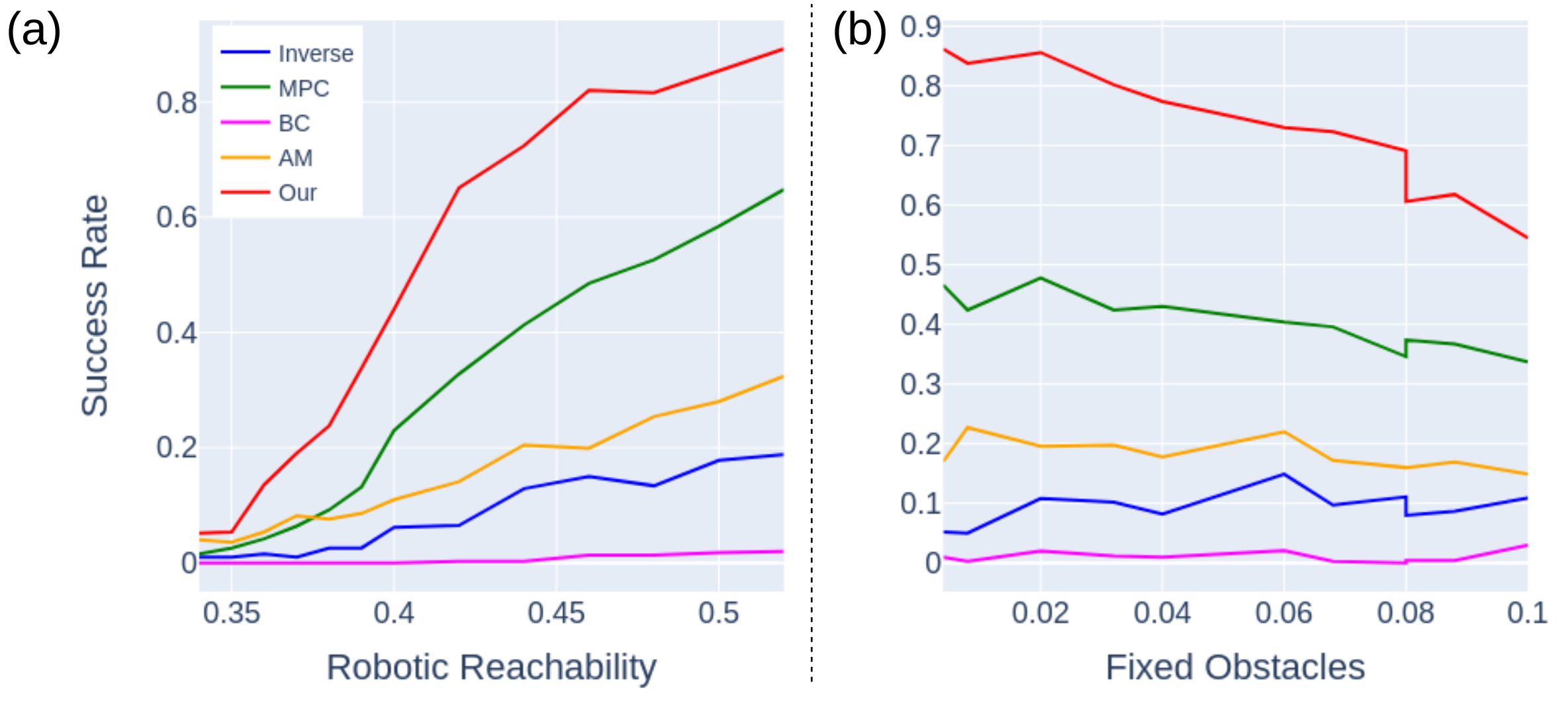"}}
    \caption{
    Comparing success rates across methods. (a) The success rate as a function of the reachability of an arm $\epsilon_f$. (b) The success rate as a function of the size of individual obstacles $\epsilon_b$.  
    }
    \label{9_curve}
\end{figure}
% In this work, we argue that learning the dynamics for deformable objects in environments with constraints such as obstacles is difficult and that we should instead learn only the unconstrained dynamics and a classifier to predict when those dynamics are valid.
The motivations of this work are: (1) Learning the dynamics of managing a deformable object with dual arms in an unstructured environment is challenging, especially when using non-fixed contact; (2) Interleaving planning and control is necessary for a long-horizon sparse reward task without a goal specification. 
% To support these claims, two kinds of 
To show the substantial improvements in our methods corresponding to the above arguments, we compare our method against various baselines. For the first claim, we compare our approach to model-based controllers, which rely on knowing the dynamics of the object.
% methods which learn the dynamics of the system to design the controllers.
% at first and then use various controllers. We learn a forward dynamic model $f(S^{[t]},A^{[t]})\approx S^{[t+1]}$ that takes the current state and the executed action as input and predicts the achieved state. 
Due to the lack of a goal specification, we also train an embedding model $f_A(S^{[t]})\approx Z^{[t]}$ to plan a goal $S^*$ in the latent space, which learns to minimize the $\mathcal{L}_2$ distance between reconstructed and actual states. Specifically, we find out the achieved embedding $Z_J^{[H+1]}$ in the dataset $\mathcal{D}$ whose $\mathcal{L}_2$ distance to the encoding $Z^{[t]}$ of the current state $S^{[t]}$ is smallest and assign the corresponding state $S_J^{[H+1]}$ as the goal $S^*$. The details of individual model-based controllers are:
% Based on these models, two controllers are design.
% design two controllers based on the dynamics models: \textbf{Inverse dynamic} and \textbf{Model Predictive Control}. 
\begin{itemize}
    \item \textbf{Inverse Dynamics}: A self-supervised goal-reaching model $\pi_I(A^{[t]}|S^{[t]},S^{*})$ to perform modeling and control \cite{ghosh2020learning}. 
    \item \textbf{Model Predictive Control}: Based on the forward dynamics model $f_D(S^{[t+1]}|S^{[t]},A^{[t]})$, a sampling-based controller is implemented to achieve one-step optimal predictive control \cite{WilsonYan2020LearningPR}. 
\end{itemize}
For the second claim, we compare two model-free techniques that attempt to shorten the distance to the goal space at each step. The details of individual model-free controllers are:
% The details of these baselines are:
\begin{itemize}
    \item \textbf{Behavioral Cloning} Based on the collected trajectories, the state is mapped to the action $\pi_{BC}(A^{[t]}|S^{[t]})$ directly in an end-to-end manner \cite{Zhang2018DeepIL}.
    \item \textbf{Action Map} Similar to FlingBot \cite{ha2021flingbot}, we predict the values of multiple pre-defined action primitives.
    Specifically, the action primitives are $8$ moving directions with constant displacement discretizing in the planar space and the value function encourages the policy to push the endpoints of the DLO closer to the corresponding arms while shifting away from the obstacles.
    % the value function is defined as minimizing the distances between the ends and the arms while maximizing the distances between the ends and the obstacles.
\end{itemize}

All the models are multi-layer perceptrons (MLP) with two hidden layers of size $256$ followed by ReLU activation functions. All the baselines share the same state $S$ provided by our representation model $f_R(S^{[t]}|I^{[t]})$, while the action is denoted as two three-dimensional vectors: the first is the index of the pick keypoint on the DLO, and the last $2$ are the $x,y$ delta direction to shape the DLO.
In order to compare the baselines fairly, all the dataset used to train the models is explored in the simulation with the same resources, whose action determination is random without human supervision.
% all the models are trained with the same dataset or a different dataset with same exploring resources. 
% For all data collection procedures, all actions are sampled randomly under systematic constraints. 
Owing to the environmental and reachability constraints, the predicted action of the models are adjusted if they are not feasible. Specifically, the search space for the index of the discrete picking points is within $[-1,1]$, while the action is discarded if a feasible one cannot be found within this range. 
In $\textbf{Action Map}$, all keypoints $Q^{[t]}=\{q_k^{[t]}\}_{k=1}^M$ are explored for an available solution.
% For the action primitives in $\textbf{Action Map}$, the algorithm seeks for an available solution within all keypoints $Q^{[t]}=\{q_k^{[t]}\}_{k=1}^M$.
% all the keypoints are search to obtain a feasible solution. 

\begin{table}[!t]
\renewcommand{\arraystretch}{1}
\footnotesize
% \caption{Comparison Results with various baselines under the planning and control framework }
% \caption{Comparisons baselines }
\caption{Quantitative comparisons between various model-based or model-free baselines}
\ADLnullwidehline
\arrayrulecolor{black}
\centering
\begin{center}
\begin{tabular}{c|ccc} 
\hline
\textbf{Policy} & Success & Mean & Std  \\ 
 & Rate \% & actions  & actions  \\ 
\hline
\textbf{Inverse} & 10.5 & 18.3 & 5.2 \\
\textbf{MPC} & 43.6 & 14.4 & 7.3  \\
\textbf{BC} & 1.0 & 19.8 & 1.5  \\ 
\textbf{Action Map} & 18.0 & 17.7 & 5.4 \\ 
\textbf{Ours} & \textbf{86.6} & \textbf{8.0} & \textbf{6.0} \\ 
\hline
\end{tabular}
\end{center}
% \begin{tablefootnote}
% \textbf{Inverse}: Inverse Dynamics; \textbf{MPC}: Model Predictive Control.
% \end{tablefootnote}
\begin{tablenotes}
\item \footnotesize{\textbf{Inverse}: Inverse Dynamics; \textbf{MPC}: Model Predictive Control; \textbf{BC}: Behavioral Cloning.}
\end{tablenotes}
\arrayrulecolor{black}
\label{table_compare_baseline}
\end{table}

We conduct multiple trials with different robotic reachability $(\epsilon_f)$ and fixed obstacle $(\epsilon_b)$ settings in order to thoroughly assess the performance of the baselines. Fig. \ref{9_curve} displays the success rates associated with various limitations throughout $500$ experiments.
% To evaluate the performance of the baselines comprehensively, we conduct several experiments under various settings of robotic reachability $(\epsilon_f)$ and fixed obstacles $(\epsilon_b)$. The success rates corresponding to different constraints across $500$ trials is shown in Fig. \ref{9_curve}. 
% illustrates the success rates of them.
% The success rates of the baselines with respect to the constraints of reachability and obstacles are shown in Fig. \ref{9_curve}. 
Our proposed algorithm reaches the highest success rate in all settings. Additional quantitative results of a specific setting $(\epsilon_f = 0.45m,\epsilon_b = 0.04m)$ are shown in Table \ref{table_compare_baseline}.
% Table. \ref{table_compare_baseline} shows the quantitative results comparing our method against baselines in simulation. 
Our method performs better in terms of three evaluation metrics, success rate, mean action and standard deviation. Without requiring human engineering programming or professional demonstration, our solution always yields satisfactory performance. 

% Due to the lack of a specific goal specification, \textbf{Inverse Dynamics} do not have a clear goal to condition.

In the following, we examine the potential causes of the aforementioned findings.
\textbf{
Inverse Dynamics} do not have a clear aim to infer since there is no particular goal definition.
\textbf{MPC} outperforms all other baselines because it is effective at reducing the cost of a long-horizon process. 
However, its performance is affected due to the inaccuracy of the forward dynamics model in this complex configuration.
% However, the correctness of the forward dynamics model in this complex configuration has an impact on its performance.
% Due to the lack of a specific goal specification, \textbf{Inverse Dynamics} do not have a clear goal to condition. Although \textbf{Model predictive control} is good at minimizing the cost of a long-horizon process, its performance is affected due to the inaccurate dynamics model in this complex configuration. 
Due to its limited generalization, \textbf{Behavioral Cloning} has the weakest performance. In addition, direct end-to-end mapping accumulates errors in the long-horizon procedure.
% huge amount of required expert demonstrations and the error accumulation in the long-horizon process. 
Although \textbf{Action Map} compresses the continuous action space by discretization, it simply concentrates on the regional maximum of the value function instead of emphasizing a long-term return. Additionally, this approach necessitates time-consuming human expertise, such as the design of action primitives and task-dependent value functions, both of which are challenging to construct for a sparse reward task in complicated configurations.

% In addition, this method requires time-consuming human knowledge, including the action discretization and a task-dependent value function, which are difficult to design for a sparse reward task under complex configurations. 
% a good value function requires engineering task-dependent knowledge, which is challenging to design for a sparse reward task under complex configurations. 
% Not only does not our method require any expert demonstration or manual engineering task programming, but also it achieves good performance in most of cases. 

\subsection{Physical Robot Demonstrations}
In this section, we show how well our suggested framework works to transfer from simulation to reality without any fine-tuning. We contend that it is advantageous to interleave planning and control for complicated manipulation tasks with limitations.
% In this section, we demonstrate how well  capability of our proposed framework from simulation to real. We argue that it is beneficial for interleaving planning and control for bimanual manipulation under environmental constraints. 
% \textbf{Hardware Systems}
\begin{figure}
\centering
    \centerline{\includegraphics[width=0.9\columnwidth]{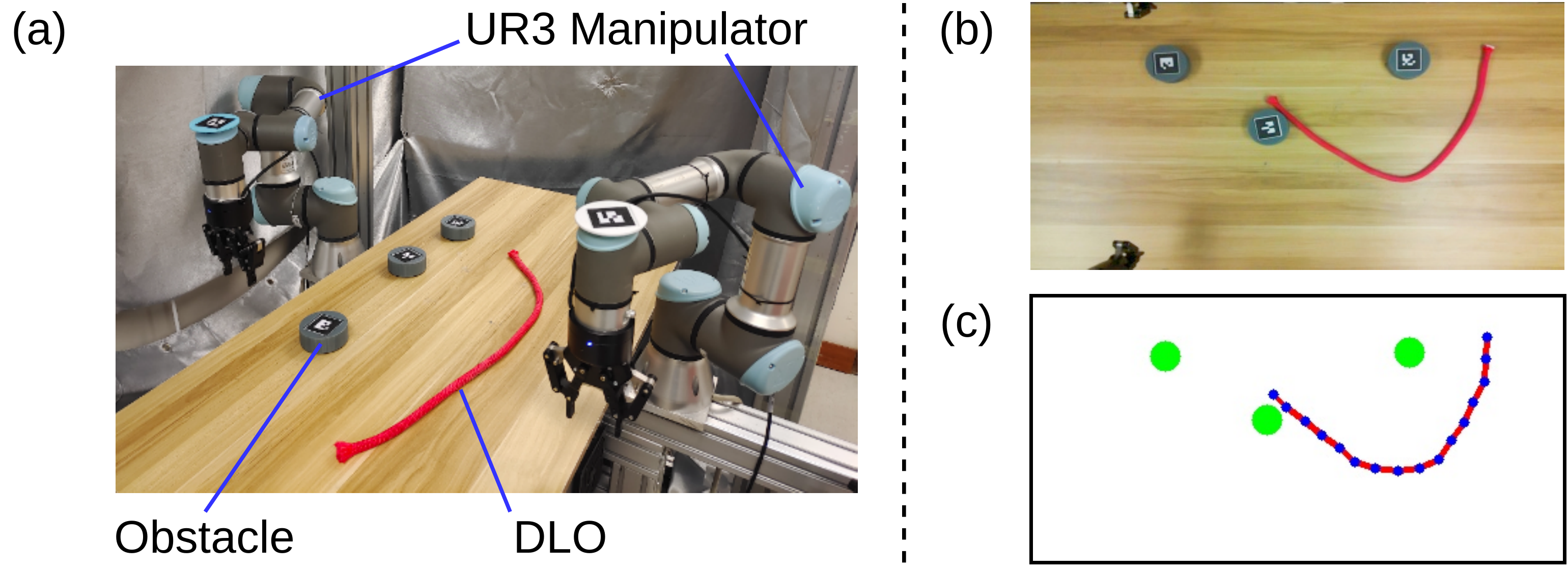}}
    \caption{The scenarios of the real environments: (a) Experimental Setup on a planar workspace, consisting of two UR3 manipulators, obstacles and a DLO. (b) Raw visual observation $I^{[t]}$ provided by the top-down camera. (c) Visualization of the extracted state $S^{[t]}$ via the representation model $f_R(S^{[t]}|I^{[t]})$. 
    }
    \label{9_real_setup}
\end{figure}

\begin{figure*}
\centering
    \centerline{\includegraphics[width=0.9\textwidth]{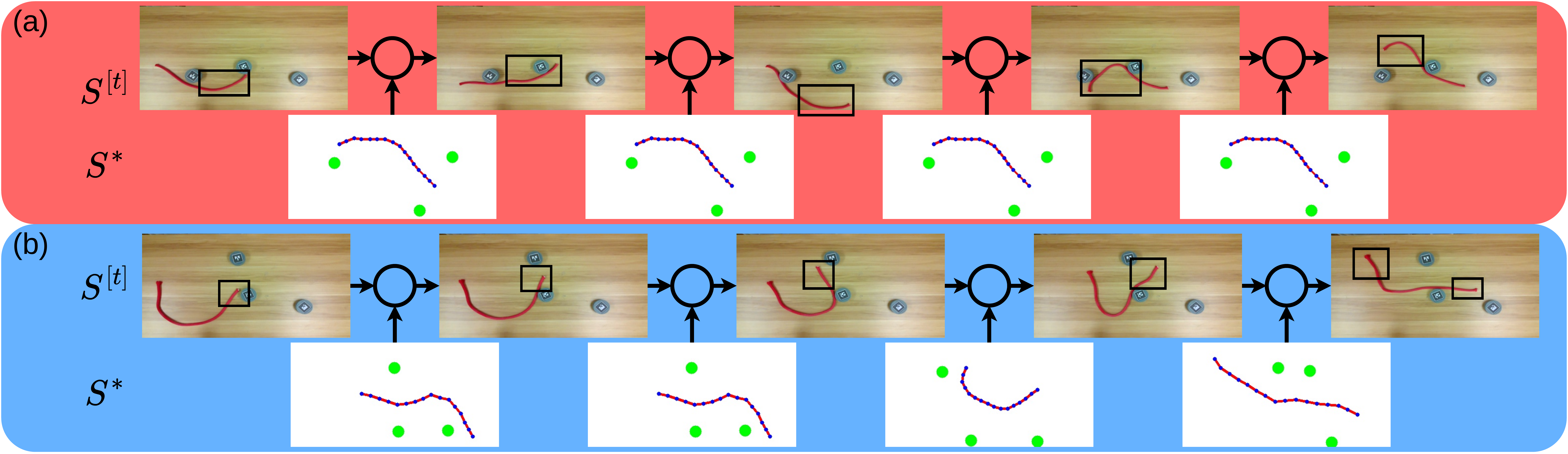}}
    \caption{Pictures of two typical examples in physical robot demonstrations. The black box highlights the manipulated region during action execution.
    % A trajectory for an episode in the simulation. At each time-step, the planning model finds the similar one in the dataset (green) with respect to the current state (red) and obtain the corresponding achieved goal (blue). 
    }
    \label{10_real_demo}
\end{figure*}

Fig. \ref{9_real_setup} shows our physical robotic environments. Two ur3 manipulators equipped with 2-fingered Robotiq grippers are used for this constrained bimanual manipulation task. The obstacles in the environment are localized with markers and fixed during an episode. An Intel Realsense L515 camera is attached to sense the top-down perspective of the environment $I^{[t]}$, as illustrated in Fig. \ref{9_real_setup}(b). Our representation model $f_R(S^{[t]}|I^{[t]})$ extracts the state $S^{[t]}$ from the raw observation $I^{[t]}$, consisting of sequential keypoints $\{q_k^{[t]}\}_{k=1}^{M}$ and fixed obstacles $\{o_b\}_{b=1}^B$, visualized in Fig. \ref{9_real_setup}(c).

We implement $100$ trials in real-world environments with a success percentage of $\textbf{90\%}$. The mean value and standard deviation of actions are $\textbf{4.48}$ and $\textbf{2.79}$ respectively. Throughout the trials, we make use of the policy model $\pi(A^{[t]}|S^{[t]},\mathcal{S}_G)$ trained in simulation and then applied it to actual situations without manual fine-tuning.
% donot require any manual fine-tuning in real experiments.
These findings illustrate that our proposed framework fills out the gap between simulation and real scenarios without any prior knowledge about the physical properties of the deformable object.
% These great performance illustrates that our algorithm extracts the temporal relationship of the task and fill out the gap from simulation to real.

To analyze our framework in detail, we provide two typical examples in the trials, visualized in Fig. \ref{10_real_demo}. Fig. \ref{10_real_demo}(a) shows an episode with a constant subgoal that is presented throughout the whole episode.
% which has the same subgoal during the whole episode.
The local goal-conditioned controller initially arranges the DLO to the center of the workspace, allowing dual arms to engage in the subsequent manipulation. The DLO is then adjusted with dual arms, namely rotating it around the obstacle. In order to bypass environmental restrictions, robots finally shift the DLO further from the obstruction.
% At last, robots move the DLO far from the obstacle to avoid environmental constraints. 
% , achieving the goal.
% Then, dual arms seek to move the left section upwards. 
We acknowledge that attached state and the planned subgoal vary in certain ways. 
Actually, rather than requesting the controller to explicitly attain a particular state, the planner is used to indicate a promising way to approach the goal space.
% there are some differences between the achieved state and the subgoal. We do not require the state reaches the subgoal $S^*$ explicitly, but it provides a promising direction for the goal-conditioned controller to determine the action.
% In other words, the planner seeks to output a most promising state within the goal space $\mathcal{S}_G$ for the current state $S^{[t]}$ to approach.
% the subgoal $S^*$ is a state within the goal space $\mathcal{S}_G$ that is easy to achieve from the current state $S^{[t]}$.
% this subgoal is suitable for dual arms to manipulate to approach the goal space $\mathcal{S}_G$ (we do not require the state reaches the subgoal explicitly). 
Owing to the replanning operation, the desired subgoal probably varies throughout the episode, as shown in Fig. \ref{10_real_demo}(b). 
% illustrates an example with varying subgoals during the episode. 
In the beginning, the controller attempts to maneuver the DLO through the barriers by moving it to the right of the workspace. A new subgoal $S^*$ is included to promote shifting the right end of the DLO to the upper right corner as the state of the DLO changes.
% At the beginning, the controller seek to move the DLO to the right of the workspace, aiming to pass through the obstacles. As the state of the DLO changes, a new subgoal $S^*$ is presented to encourage moving the right end of the DLO to the upper right corner. 
Then, both arms participate in distributing the DLO horizontally in the workspace based on a new subgoal $S^*$.
% After moving the right end of the DLO away from the obstacle, another new subgoal is obtained to distribute the DLO flatly in the workspace. At this step, dual arms manipulate together to deform the DLO globally. 
% This example illustrates that replanning is necessary in this challenging constrained bimanual manipulation task, which our algorithm performs good to provide different promising directions for the goal-conditioned controllers at different stages.
This example illustrates that replanning is useful to adjust the reaching direction in this challenging constrained bimanual manipulation task.

% rearranges the middle part of the DLO to make the right part of the DLO leave the obstacles. Then, one arm is able to grasp the right part of the DLO to approach the new subgoal $S^*$. Finally, dual arms are assigned actions to shape the relevant regions around the corresponding ends to a state within the goal space $\mathcal{S}_G$.

\begin{figure}
\centering
    \centerline{\includegraphics[width=\columnwidth]{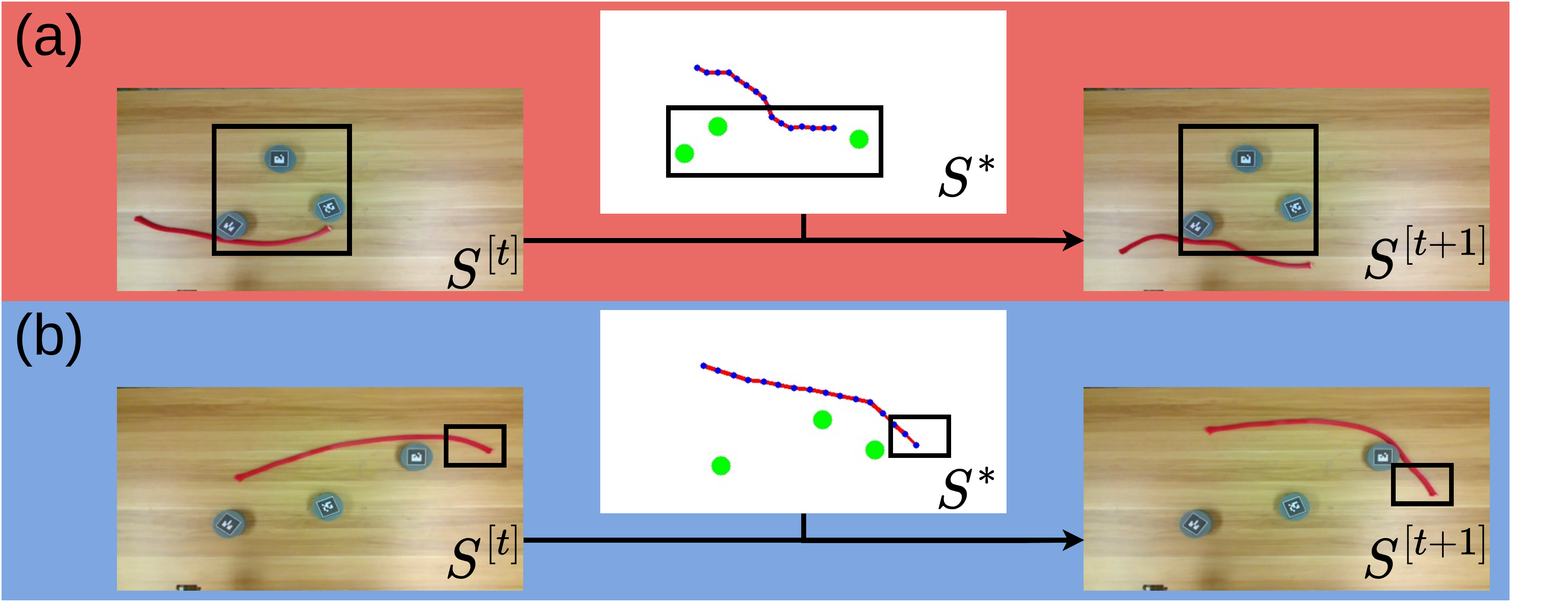}}
    \caption{Pictures of two failed examples in physical robot demonstrations. The black box highlights the focused region when analyzing the cause of the failures.
    }
    \label{11_real_fail}
\end{figure}

Although our framework is capable of handling the majority of the challenging tasks, there are some situations when it fails. Fig. \ref{11_real_fail} presents two typical failure examples. The failure case in Fig. \ref{11_real_fail}(a) is mainly caused by the planner. Our contrastive learning-based planner is driven by a desire to investigate the temporal information about the relative distribution between the DLO and the obstacles in the successful experiences. However, the provided subgoal $S^*$ is not appropriate for the present state to pursue.
% in this episode fails to provide a promising direction for the current state.
Specifically, the obstacles of the episode are in the middle of the workspace, which is a conflict with the subgoal (the obstacles distribute on the side and the major part of the DLO distribute in the middle). 
% This unsuitable goal makes the local goal-conditioned controller hard to align. 
The possible reason for this phenomenon is our contrastive learning-based encoder $f_E(Z|S)$ incorrectly classifies certain related states in the latent space. Another failure case in Fig. \ref{11_real_fail}(b) is mainly caused by the controller. The leader-follower control scheme selects the points around the right end to approach the subgoal $S^*$. However, this implementation makes the corresponding part of the DLO out of reachability of dual arms, thus failing to manipulate the interested region (around the right end) towards the subgoal $S^*$ further.
% Moving far from the dual arms $\{a_r\}_{r=1}^2$, the controller is not capable of manipulating the intersted region (around the right end) towards the subgoal $S^*$ further. 
% As a result, dual arms move the DLO to the left, making the DLO return to a state similar to previous situation. 
% As a result, only other alignment areas are accessible for execution, returning the DLO to a state similar to previous situation.
As a result, only other areas are accessible for the correspondence-based manipulation, returning the DLO to a state similar to the previous situation.
% Due to the systematic constraints of reachability and obstacles, dual arms can not grasp the region around the ends towards the subgoal $S^*$ further. 
% As a result, dual arms can only manipulate the middle region of the DLO, making the DLO returned to a state similar to the previous one.
The controller recursively performs the two aforementioned types of actions, namely trapping in this local matching discrepancy.

\section{Conclusion}
In this paper, we propose a novel framework for dexterous bimanual manipulation under environmental and reachability constraints. Removing the assumption of object rigidity and a goal specification, our proposed methodology further enhances the intelligence of bimanual manipulation. To deal with the long-horizon complexity, our policy model is factorized into global subgoal planning and local goal-conditioned control. 
% Interleaving global planning and local control, the policy model outputs the action to manipulate the object towards the goal space along a promising and feasible direction.
% Due to the lack of goal specifications in the long-horizon sparse reward setting, we plan a feasible but also promising subgoal. 
% Based on the collected dataset in simulation, we learn an embedding model via contrastive learning that extract the spatial-temporal information of successful episodes. 
Our subgoal planner provides a promising direction for the state in the query to pursue through similarity matching in the embedding space, which is encoded by an encoder trained in a contrastive learning manner. 
Our controller leverages a leader-follower scheme to determine the collaborative correspondence-based action of dual arms directed by the subgoal.
% Conditioning on the planned subgoal, our local controller implements a correspondence-based alignment between the current state and the desired subgoal. A leader-follower control scheme is leveraged to determine the collaboration between dual arms under environmental and reachability constraints.
% systematic constraints of reachability and obstacles. 
All the models are trained in simulation and can be transferred to real environments without any fine-tuning. A detailed experimental study is reported to illustrate the effectiveness of the framework.
% To validate the performance of our new framework, we reported a detailed experimental study and demonstrate the effectiveness in real-world.  

However, our methods exhibit some limitations. We choose the state that is closest to the query during each planning, while the uncertainty can not be evaluated. In some situations, the correspondence-based controller traps in a local minimum. For future directions, we are interested to estimate the utility of the subgoal in planning and a feedback-based predictive controller.

% \section{References}
% You can use a bibliography generated by BibTeX as a .bbl file.
%  BibTeX documentation can be easily obtained at:
%  http://mirror.ctan.org/biblio/bibtex/contrib/doc/
%  The IEEEtran BibTeX style support page is:
%  http://www.michaelshell.org/tex/ieeetran/bibtex/
% You can manually copy in the resultant .bbl file and set second argument of $\backslash${\tt{begin}} to the number of references
%  (used to reserve space for the reference number labels box).
\bibliographystyle{ieeetr}
\bibliography{ref}

\end{document}